\documentclass{article}

\usepackage{arxiv}

\usepackage[utf8]{inputenc} 
\usepackage[T1]{fontenc}    
\usepackage{hyperref}       
\usepackage{url}            
\usepackage{booktabs}       
\usepackage{amsfonts}       
\usepackage{nicefrac}       
\usepackage{microtype}      
\usepackage{lipsum}         
\usepackage{graphicx}
\usepackage{doi}
\usepackage{graphicx}
\usepackage{float}
\usepackage{subfigure}
\usepackage{epstopdf}
\usepackage[dvips]{epsfig}
\usepackage{lineno,hyperref}
\usepackage{anysize}
\marginsize{3cm}{3cm}{2cm}{2cm}
\usepackage{amscd}
\usepackage{mathrsfs}
\usepackage{verbatim}
\usepackage{graphicx}
\usepackage{amsmath}
\usepackage{cleveref}       
\usepackage{amssymb}
\usepackage{array}
\usepackage{booktabs}
\usepackage{soul,xcolor}

\usepackage{soul}  
\usepackage{color}
\usepackage{lineno}
\setstcolor{red}

\usepackage{adjustbox}
\usepackage{mdframed}
\usepackage{multicol}
\usepackage{multirow}
\usepackage{algorithm}
\usepackage{algpseudocode}

\title{Contrastive Multi-Modal Representation Learning for Spark Plug Fault Diagnosis}

\newif\ifuniqueAffiliation
\uniqueAffiliationtrue

\ifuniqueAffiliation 
\author{Ardavan Modarres \\
	Department of Electrical Engineering\\
	K. N. Toosi University of Technology\\
	Tehran, Iran \\
	\texttt{ardavan.modarres@email.kntu.ac.ir} \\
	\And
	Vahid Mohammad-Zadeh Eivaghi \\
	Department of Electrical Engineering\\
	K. N. Toosi University of Technology\\
	Tehran, Iran \\
	\texttt{vmohammadzadeh@email.kntu.ac.ir} \\
	\And
	Mahdi Aliyari Shoorehdeli \\
	Faculty of Electrical Engineering\\
	K. N. Toosi University of Technology\\
	Tehran, Iran \\
	\texttt{aliyari@kntu.ac.ir} \\
	\And
	Ashkan Moosavian \\
	Department of Agricultural Engineering\\
	Technical and Vocational University\\
	Tehran, Iran \\
	\texttt{a\_moosavian@tvu.ac.ir} \\
}
\else
\usepackage{authblk}

\newbox{\orcid}\sbox{\orcid}{\includegraphics[scale=0.06]{orcid.pdf}} 
\author[1]{%
	\href{https://orcid.org/0000-0000-0000-0000}{\usebox{\orcid}\hspace{1mm}David S.~Hippocampus\thanks{\texttt{hippo@cs.cranberry-lemon.edu}}}%
}
\author[1,2]{%
	\href{https://orcid.org/0000-0000-0000-0000}{\usebox{\orcid}\hspace{1mm}Elias D.~Striatum\thanks{\texttt{stariate@ee.mount-sheikh.edu}}}%
}
\affil[1]{Department of Computer Science, Cranberry-Lemon University, Pittsburgh, PA 15213}
\affil[2]{Department of Electrical Engineering, Mount-Sheikh University, Santa Narimana, Levand}
\fi

\hypersetup{
pdftitle={Contrastive Multi-Modal Representation Learning for Spark Plug Fault Diagnosis},
pdfsubject={Multi-Modal Deep Learning, Deep Learning, Multi-View Learning, Multi-Modal Data Fusion, Machine Learning, Intelligent Condition Monitoring, Fault Diagnosis},
pdfauthor={Ardavan Modarres, Mahdi Aliyari, Vahid Mohammad-Zadeh Eivaghi, Ashkan Moosavian},
pdfkeywords={Multi-Modal deep learning, Multi-View learning, Multi-Modal data fusion, fault diagnosis, internal combustion engine, spark plug, vibration signals, acoustic signals},
}

\begin{document}
\maketitle

\begin{abstract}
Due to the incapability of one sensory measurement to provide enough information for condition monitoring of some complex engineered industrial mechanisms and also for overcoming the misleading noise of a single sensor, multiple sensors are installed to improve the condition monitoring of some industrial equipment. Therefore, an efficient data fusion strategy is demanded. In this research, we presented a Denoising Multi-Modal Autoencoder with a unique training strategy based on contrastive learning paradigm, both being utilized for the first time in the machine health monitoring realm. The presented approach, which leverages the merits of both supervised and unsupervised learning, not only achieves excellent performance in fusing multiple modalities (or views) of data into an enriched common representation but also takes data fusion to the next level wherein one of the views can be omitted during inference time with very slight performance reduction, or even without any reduction at all. The presented methodology enables multi-modal fault diagnosis systems to perform more robustly in case of sensor failure occurrence, and one can also intentionally omit one of the sensors (the more expensive one) in order to build a more cost-effective condition monitoring system without sacrificing performance for practical purposes. The effectiveness of the presented methodology is examined on a real-world private multi-modal dataset gathered under non-laboratory conditions from a complex engineered mechanism, an inline four-stroke spark-ignition engine, aiming for spark plug fault diagnosis. This dataset, which contains the accelerometer and acoustic signals as two modalities, has a very slight amount of fault, and achieving good performance on such a dataset promises that the presented method can perform well on other equipment as well.
\end{abstract}

\section{Introduction}      \label{section.intro}
\paragraph{}
Internal combustion (IC) engines have been one of the most important types of equipment for power generation for a wide range of machineries consisting of many complex engineered mechanisms. Spark-ignition (SI) engine is a class of IC engines. The spark plug is one of the most important components of an SI engine which is responsible for producing an appropriate ignition. The spark plugs have a high failure potential in all engines. Some of the spark plug faults are listed in  \cite{1}. Faults related to spark plug gap are very attention-receiving among various types of spark plug defects due to their relation to engine combustion quality \cite{2}. 
The most common defect in spark plugs is the growth of spark plug gap due to engine function, which may be caused by corrosion, erosion, oxidation, rich fuel-air mixture, reaction with gasoline's lead, etc. By increasing the voltage required to generate appropriate sparks between electrodes more than normal conditions, the increased gap causes damage to the ignition system, and it has undesirable effects on both the spark plug and engine performance, as well as the engine's electrical system. The aforementioned defect can postpone the spark between electrodes and may result in pre-ignition, which is one of the reasons for the misfire and knocking phenomenon \cite{3}. Moreover, the increased spark plug gap is one of the most important reasons for air pollution, combustion quality reduction, and a decline in engine performance. Hence, recognizing the increased spark plug gap is crucial for preventing performance reduction and further collateral damages in an SI engine. As a result, condition monitoring and fault diagnosis of spark plugs play a vital role in maintaining SI engines performance and increasing their remaining useful life.
\paragraph{}
Since most faults, including spark plug gap growth, change the vibration and acoustic behaviors of the engine, vibration signal analysis and acoustic signal analysis are two popular techniques that show promising results. Vibration-based analysis methods utilize vibration signals measured from the equipment and have the advantage of low cost and rapid measurement \cite{2}. Acoustic-based analysis fault diagnosis approaches are non-invasive since the microphone sensor is non-contact and can be used when it is not possible to mount an accelerometer sensor on the equipment and acquire sound signals at a distance from the equipment \cite{4}. 
\paragraph{}
Data-driven machine health monitoring and condition monitoring techniques in industrial manufacturing have been revolutionized by the development of sensors and data provided by the subsequent widespread utilization of low-cost sensors. Using machine learning techniques in data-driven condition monitoring approaches allows us to make appropriate decisions from vast amounts of data. The conventional machine learning techniques used in data-driven condition monitoring include hand-crafted feature extraction, feature selection, and shallow decision-making model training. The development of the most attention-receiving and fastest-growing subfield of machine learning, deep learning \cite{5}, provides an excellent opportunity for big data processing and analysis showing promising achievements beyond mankind's capability, and dramatically improved performance in visual object recognition, object detection, automatic speech recognition, natural language processing, and drug discovery \cite{6}. Compared with traditional machine learning methods, deep learning is capable of yielding informative features from original raw signals through a series of hierarchical transformations followed by a nonlinear activation function and is trained in an end-to-end fashion without human labor-intensive intervention and expert knowledge for manual feature extraction. The features extracted by deep neural networks can ultimately be beneficial for a classification or regression downstream task. As a result, deep learning is considered an attractive option to process industrial big data and has been widely used in machine health state classification and fault diagnosis. 
\paragraph{}
Convolutional Neural Network (CNN) \cite{7} is a typical deep learning scheme widely used in different pattern recognition problems. In recent years, scholars have tried to use CNNs to process vibration and acoustic signals of mechanical systems in order to develop high-performance equipment health state classification solutions. In \cite{8}, a one-dimensional CNN (1DCNN) architecture is proposed for processing raw vibration signals for gearbox fault diagnosis. In \cite{9}, time-frequency features are extracted from raw vibration signals using wavelet transform and fed into a 2DCNN for gearbox fault diagnosis. In \cite{10}, an adaptive deep convolutional neural network is proposed for processing acoustic emission signals and bearing health state classification. In \cite{11}, an effective and reliable method based on CNN and discrete wavelet transformation (DWT) is proposed to identify the fault conditions of wind turbine planetary gearboxes. In \cite{12}, an Enhanced Convolutional Neural Network (ECNN) with enlarged receptive fields is proposed for intelligent health state identification of planetary gearboxes. In \cite{13}, a CNN architecture is proposed for rotating machinery fault recognition. Another well-performing neural network scheme that has been widely utilized in different applications, including machine health monitoring, is the Autoencoder \cite{14}. In \cite{15}, selective stacked denoising Autoencoders are utilized to extract meaningful features from raw vibration signals for gearbox fault diagnosis. 
\paragraph{}
Despite the achievements of the above-mentioned works, they are based on only one sensory measurement (one modality) and only use one of the vibration or acoustic modalities to perform health state classification. Because of the complex nature of many industrial mechanisms, fault diagnosis based on one modality measurement often fails to take the sophisticated fault nature into account, and it can be misguided by single modality noise. Thus, an efficient data fusion strategy is demanded to combine information provided by multiple modalities of sensory measurement into an enriched common representation in order to improve the performance of equipment health state monitoring and decrease the missed detection rate. 
\paragraph{}
Most traditional data fusion approaches use manually extracted statistical features and combine these multi-modal features into a long vector by simply concatenating them to achieve data fusion. Another great performing approach for taking advantage of multi-modal measurements is the Multi-View Learning (MVL) \cite{16} paradigm. 
\paragraph{}
Canonical Correlation Analysis (CCA) can be regarded as the starting point of the Multi-View Learning paradigm developed in \cite{17, 18} at first for processing two sets of random variables (modalities) jointly, which seeks a linear transformation for each view, such that after applying the transformations, the projected modalities in the new subspace are maximally correlated. As a result, multiple modalities are mapped to an information-enriched common representation. Various extensions of CCA, such as kernel and nonlinear CCA \cite{19}, locality preserving CCA \cite{20}, etc., have been developed. Some researchers proposed deep versions of CCA in combination with deep neural networks \cite{21}, and apart from deep CCA, some proposed approaches entirely based on neural networks to tackle common representation learning problem \cite{22, 23}. Some researchers leveraged the merits of both CCA and Autoencoders to project multiple input modalities into a common maximally correlated subspace; for instance, in \cite{24}, the Correlational Neural Networks (CorrNet), which is an Autoencoder-based architecture with CCA advantages is introduced. Multi-modal learning has been widely utilized in many applications, from lung nodule classification \cite{25} in the medical domain to 3D object recognition \cite{26}. 
\paragraph{}
In the current research, we utilized the vibration and acoustic signals and presented a multi-modal Autoencoder based on Multi-View Learning paradigm with a unique training procedure for data fusion aiming for fault diagnosis applications, which is evaluated by engine spark plug gap fault diagnosis case study. To the best of our knowledge, the unique training procedure is being introduced in the fault diagnosis domain for the first time. In \cite{27}, it is shown that the fusion of vibration and acoustic signals improves the diagnosis of the increased gap fault of the spark plug in the case study utilized in current research. The unique training procedure introduced in the current research not only proves to be a great end-to-end data fusion strategy and succeeds in combining multiple modalities of sensory measurements but also enables our proposed neural network to utilize only one of the acoustic or vibration signals during the test time while still achieving similar performance as using both, which results in a more robust and cost-effective condition monitoring system. The above-mentioned achievement is the main novelty of this paper. 
\paragraph{}
Fewer studies in the field of condition monitoring are focused on spark plug fault recognition. In \cite{28}, the instantaneous angular speed method is used for automobile engine fault diagnosis. In this research, the spark plug gap was increased by $2$ mm for the simulation of a faulty condition. This increase in the spark plug gap results in a severe fault, which is easy to recognize and may not occur in practice. In \cite{29}, engine fault diagnosis is investigated in which the spark plug gap increased from $1$ mm to over $2$ mm (i.e., more than $100\%$ increase in the spark plug gap). This amount of spark plug gap which is almost the same as the one in \cite{28} may not occur in practice. Moreover, the considered fault is applied on the spark plug of all cylinders simultaneously for simulation of a faulty scenario, which results in severe fault and is easy to identify and may not occur in practice. In the present study, a much slighter degree of the aforementioned fault is investigated in which the spark plug gap growth is less than $60\%$ percent for simulating faulty scenarios, which is much more probable to occur in practice and difficult to recognize. Also, the considered fault is applied on one cylinder at a time, which leads to a harder-to-recognize slight fault with more practical frequency of occurrence. The two-above mentioned considerations are two other novelties of the present study. In conclusion, our contributions are:
\begin{itemize}
\item We present a new methodology based on Multi-View Learning with a unique training strategy for the first time in the fault diagnosis realm, especially on engine condition monitoring and spark plug fault diagnosis.
\item Our presented methodology takes data fusion to the next level, in which one input modality can be omitted during inference time with slight performance reduction or none at all. This capability enables the multi-modal fault diagnosis system developed based on current research methodology to achieve higher robustness in case of sensor failure occurrence during test time. Also, one can deliberately omit one of the input modalities (usually the more expensive one or the one with complicated installation) in order to achieve a cost-effective or an easy-to-use product.
\item The case study for evaluating our presented methodology is a very complex engineered mechanism, in which for faulty case simulation and data acquisition, a much slighter fault scenario is simulated compared to other research focused on spark plug gap fault diagnosis, which occurs more frequently in practice and achieving the above capability (our second contribution) on such a complicated case study make our presented methodology even more valuable.
\item The dataset of the case study is acquired under non-laboratory conditions, which makes fault diagnosis much harder and corresponds with real-world scenarios.
\end{itemize}
To the best of our knowledge, our second contribution, which is our main contribution, is being investigated for the first time in the data fusion research line, especially in multi-modal fault diagnosis.
The rest of this article is organized as follows: in section \ref{section.litreview} we review the related research to ours with focus on multi-modal machine health monitoring systems. In section \ref{section.es} we explain the system under study, data acquisition, and dataset in detail. In section \ref{section.pm} we explain our proposed method and corresponding training procedure. In section \ref{section.randd}, we present the result of the simulations. In section \ref{section.conclusion}, a brief review of the main advantages of our proposed method over related research, conclusion, and future ideas are presented.
\section{Related Works} \label{section.litreview}    \label{section.litreview}
\paragraph{}
Using multiple modalities in machine health monitoring systems is a popular approach and has attracted lots of attention. In \cite{30}, hand-crafted extracted features are concatenated and fed into an Autoencoder for fusing features for bearing fault diagnosis. In \cite{31}, time, frequency, and time-frequency hand-crafted features are extracted and fused using an Autoencoder for bearing fault identification. In \cite{32}, hand-crafted features from multiple vibration sensors are concatenated and fused using a Deep Belief Network (DBN) for bearing fault diagnosis. The above-mentioned works do not lie under the umbrella of multi-modal fault diagnosis, and they belong to the multi-sensor fault diagnosis research line since only one modality of sensory measurement (only vibration) is utilized from multiple sensors; however, one can carry out these algorithms using different modalities. The disadvantages of the above-mentioned works are using only one modality of sensory measurement and hand-crafted feature extraction, which lead to low-performance condition monitoring systems. 
\paragraph{}
In \cite{33}, a DBN is used for feature extraction from each of the acoustic and vibration signals. Then, the concatenated features are fed into a Random Forest for bearing fault diagnosis. In \cite{34}, raw time series are transformed into the time-frequency domain using wavelet transform, and a 1DCNN and a 2DCNN are used for feature extraction from raw time series and time-frequency domain representation, respectively. Finally, a Support Vector Machine (SVM) is used for bearing fault diagnosis based on fused features. In \cite{35}, a 1DCNN and a 2DCNN are utilized to extract features from one-dimensional time series and its time-frequency representation, respectively. Then, the concatenation of extracted features is fed into an SVM for bearing fault diagnosis. The main disadvantage of \cite{33, 34, 35} is utilizing conventional machine learning classification methods such as random forest and SVM, which show lower performance in comparison with neural networks. Also, in \cite{34, 35}, only one modality sensory measurement in multiple domains is utilized. Hence \cite{34, 35} can not be considered multi-modal, but one can adopt these algorithms and utilize multiple modalities. 
\paragraph{}
In \cite{36}, two 1DCNN architectures are utilized to extract features from vibration and acoustic sensors, and a Multi-Layer Perceptron (MLP) is used for fusing features from different modalities for bearing fault diagnosis. In \cite{37}, raw vibration and electric current time series of an electric motor are transformed to the time-frequency domain using Short Time Fourier Transform (STFT), and a 1DCNN is used for feature extraction from raw time series input, and a 2DCNN is used for feature extraction from the time-frequency domain input and the concatenated features from different modalities are fed into an MLP classifier for electric motor fault diagnosis. In \cite{38}, multi-modal time series are considered as different input channels and are combined using a 1DCNN. The output of 1DCNN is used for bearing fault diagnosis. In \cite{39}, RGB and Infrared (IR) images from an unmanned aerial vehicle are fused through 2DCNNs for power line equipment fault diagnosis. In \cite{40}, 1DCNNs with dynamic routing are used for feature extraction from vibration and electric current times series for electric motor fault diagnosis. In \cite{41}, an MLP network and a 1DCNN network are used to extract features from tabular data and time series, respectively, and the concatenation of these extracted features is used for gear fault diagnosis. In \cite{42}, a 1DCNN and an MLP are utilized to extract features from vibration and current sensors, and extracted features from different modalities are combined using an MLP and used for motor fault diagnosis. In \cite{43}, 2DCNNs are used for feature extraction from different input modalities, and the extracted features are fused using an MLP for bearing fault identification. In \cite{44}, a hybrid model consisting of one CNN architecture and one Long Short-Term Memory (LSTM) architecture is utilized for feature extraction from multiple modalities for Remaining Useful Life (RUL) prediction. In \cite{45}, a pre-trained VGG \cite{46} is used for feature extraction from thermal images, and an MLP network is used for feature extraction from tabular power, current, and temperature data. Then, an Autoencoder is trained on the concatenation of extracted features for fusing modality-specific features. Finally, fused features are used for industrial refrigerator fault diagnosis. In \cite{47}, a deep twin convolutional neural network with multi-domain inputs (DTCNNMI) is proposed for strongly noisy diesel engine misfire detection. In \cite{48}, vibration and acoustic signals are fused in three levels: signal level, feature level, and classifier level, using wavelet transform and artificial neural networks. The main disadvantage of the above-mentioned works is that they only leverage supervised training strategy and all the input modalities always have to be present during the test time. 
\paragraph{}
An Autoencoder is trained on the concatenated signals from different sensors for data fusion and bearing condition monitoring in \cite{49}. In \cite{50}, two DBNs are used for feature extraction from each of the electric current and vibration signals independently, and a DBN is trained on the concatenation of modality-specific extracted features in order to fuse these features for gearbox fault diagnosis. In \cite{51}, two Autoencoders are utilized to extract features from the vibration and acoustic signals of rotating machinery, and a similarity metric between the latent representation of these Autoencoders is maximized as a data fusion strategy. In \cite{52}, a multi-modal Autoencoder is utilized for gas turbine anomaly detection. The notable disadvantage of the above-mentioned Autoencoder-based data fusion strategies is taking advantage of only unsupervised learning strategy. Another noteworthy disadvantage is that all the input modalities always have to be present in both training and test time, which leads to additional costs for sensors and computational devices. 
\paragraph{}
In the current research, we take data fusion to the next level by introducing the missing modality case, in which one of the input modalities can be omitted during the test time with a slight sacrifice in performance or without any. To the best of our knowledge, the missing modality case is being introduced for the first time in the fault diagnosis domain. 
\paragraph{}
In \cite{53}, Gaussian-Bernoulli Deep Boltzmann Machines (GDBM) are used to extract features from multiple modalities, and a multi-modal SVM is used for the fault diagnosis task for a gearbox. The notable weak point of \cite{53} is that learning a discrete latent variable results in information loss, while our proposed methodology projects continuous multi-modal sensory measurements into a continuous space and does not lead to any information loss. 
\paragraph{}
In \cite{54}, classifiers decisions on acoustic and vibration sensors are fused using the Dempster-Shafer theorem for planetary gears fault diagnosis. In \cite{27}, vibration and acoustic sensors are fused using the Dempster-Shafer theorem for spark plug fault recognition. The main disadvantages of \cite{54, 27} are the lack of scalability and subjectivity in assigning belief functions besides the necessity of the presence of all input modalities during test time.
\section{Experimental Setup}   \label{section.es}
\begin{figure}[h]
	\centering
	\includegraphics[height=0.32\textwidth]{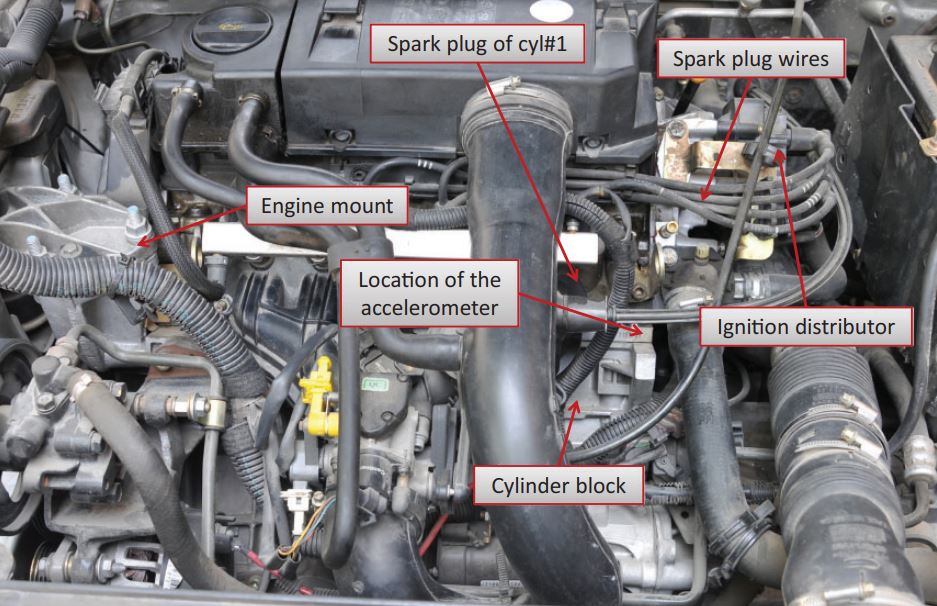}\caption{The engine under study \cite{2}}\label{fig1.exp.eng}
	\centering
\end{figure}

\paragraph{}	
In current research, the experiments were carried out on an inline four-cylinder, four-stroke SI engine. The engine under study is depicted in Figure \ref{fig1.exp.eng}.

\begin{figure}
	\centering
	\includegraphics[height=0.3\textwidth]{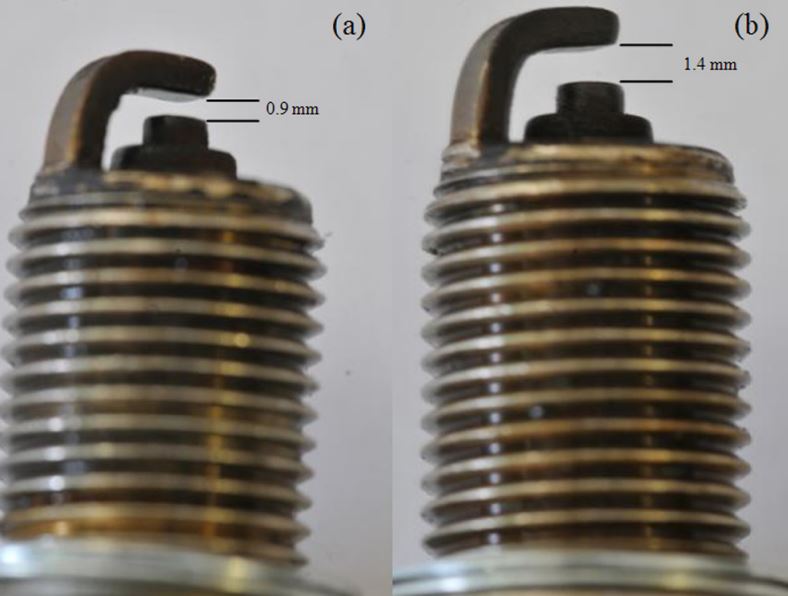}\caption{The utilized spark plugs \cite{27}: (a) healthy and (b) defective}\label{fig2.exp.plug}
	\centering
\end{figure}

To this end, healthy and defective spark plugs were installed on the engine, and vibration and acoustic signals were measured for each engine condition. In normal condition, four healthy spark plugs with a standard gap of $0.9$ mm were installed on cylinders. In order to simulate an engine's abnormal condition, a faulty spark plug with a gap of $1.4$ mm was installed. Due to spark ignition between two electrodes, gradual erosion is made in the electrodes, and consequently, the gap is enlarged, and the spark plugs may become faulty and useless \cite{2}. This fault, which usually occurs in SI engines, is also known as the wide spark plug gap. Healthy and defective spark plugs used in our experiments are depicted in Figure \ref{fig2.exp.plug}. In the present study, the three following defective conditions were considered.\\
1-Faulty spark plug was installed on cylinder $1$\\
2-Faulty spark plug was installed on cylinder $2$\\
3-Faulty spark plug was installed on cylinder. $3$\\
The four engine conditions investigated in the current research are depicted in Table \ref{table.one}.

\begin{table}[h!]
	\centering
	\caption{Different engine conditions}
	\label{table.one}
	\begin{tabular}{|c|c|c|c|c|}
		\hline
		Engine condition & Defect location & gap(mm) & Abbreviation & Number of Samples \\
		\hline
		Healthy & Normal & 0.9 & H & 244 \\
		\hline
		Defective 1 & cylinder 1 & 1.4 & C1 & 120 \\
		\hline
		Defective 2 & cylinder 2 & 1.4 & C2 & 173 \\
		\hline
		Defective 3 & cylinder 3 & 1.4 & C3 & 168 \\
		\hline
	\end{tabular}
\end{table}

\begin{figure*}[h!]
	\centering
	\includegraphics[height=0.37\textwidth]{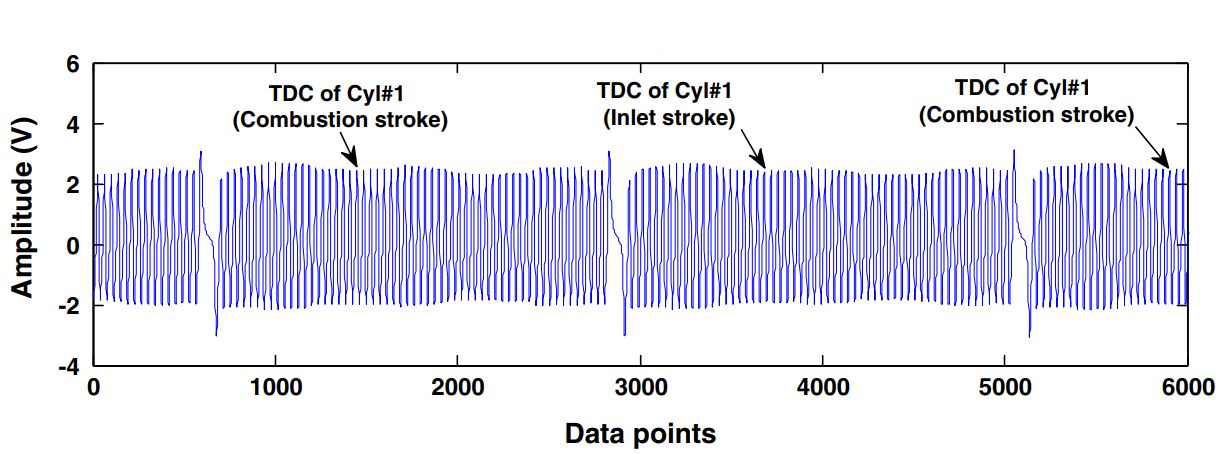}\caption{The crankshaft position signal \cite{27}}\label{fig3.exp.crankshaft}
	\centering
\end{figure*}

The engine ran on constant rotational speed ($867$ rpm) under no-load condition. The rotational speed was measured by a crankshaft position sensor. A sample signal of the crankshaft position is depicted in Figure \ref{fig3.exp.crankshaft}. The reason behind selecting the above-mentioned operating condition is that the engine runs smoothly at idle speed and no load condition, so defects may hardly appear, and fault detection becomes more difficult. 

In this case, the intelligent diagnostic system must be more powerful and precise.
Moreover, it is desired that engine health monitoring is performed under the easiest engine operating condition in practice \cite{2}.

\begin{figure}
	\centering
	\subfigure[Acoustic modality]{\includegraphics[width=0.45\textwidth]{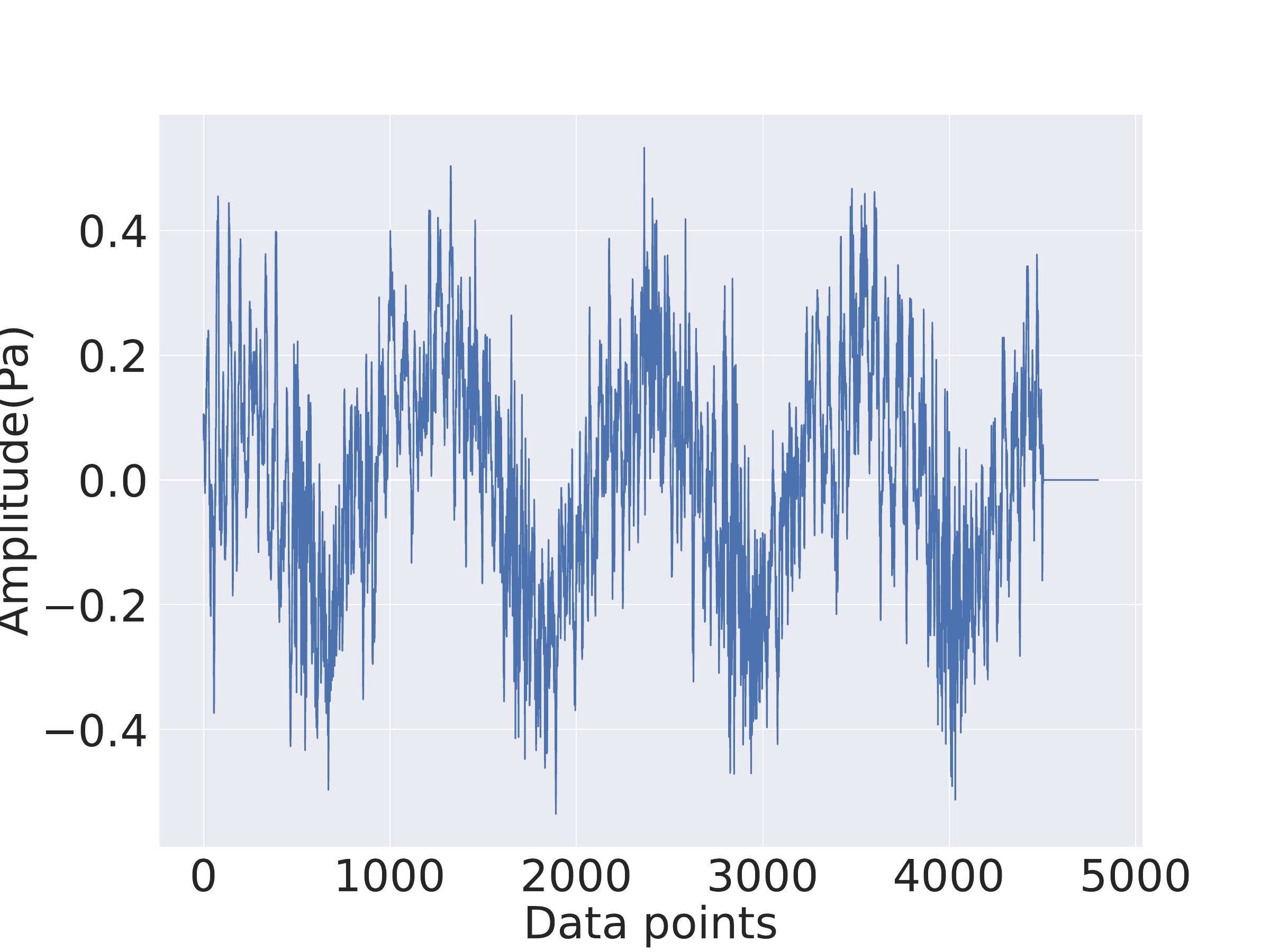}}
	\hfill
	\subfigure[Vibration modality]{\includegraphics[width=0.45\textwidth]{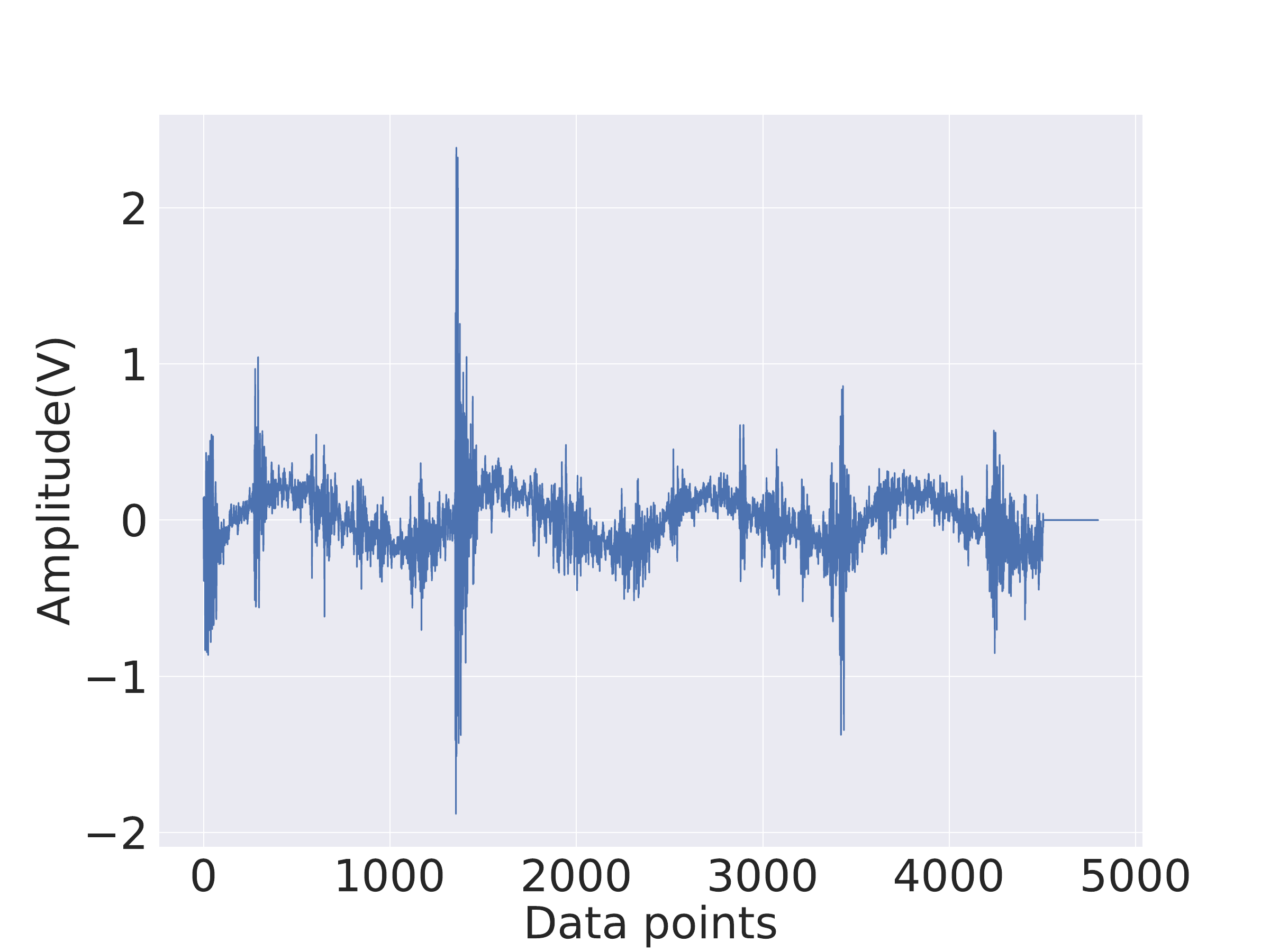}}
	\caption{A typical sample consisting of 2 modalities}\label{fig4.views}
\end{figure}

\paragraph{}
In order to collect vibration signals from the engine, an accelerometer sensor type 357B11 was mounted horizontally (transversely) in the vicinity of cylinder 1 since measurements in the horizontal (transverse) direction provide more helpful information about the distribution of harmonic components. The accelerometer has a resonant frequency of 50 kHz, which is sufficiently far away from the bandwidth of this work. In order to collect the acoustic signals emitted by the engine, a B\&K type 4188 microphone was placed on top of it. The data collection system consisted of a Brüel \& Kjær (B \& K) NEXUS conditioning amplifier type 2692 low noise version with a high bandwidth and a 17-channel data acquisition system. In the present study, the acoustic, vibration, and crank position signals were acquired using three channels. The signals were sampled simultaneously at 32,768 Hz sampling rate for five seconds. Several repeats in sampling were performed for each engine condition. The PULSE time data recorder type 7708 was used to record time domain signals, which were then transferred to a computer. 
\paragraph{}
The original captured signals had a length of 163,840 (5 s $\times$ 32,768 Hz). Each original signal was segmented into several sample signals according to the crankshaft position signal. Note that a ‘‘sample signal’’ is defined as a segmented signal with the length of two complete revolutions of the crankshaft. There are four ignitions in two complete revolutions, so each plug sparks once during the sample signal period. Therefore, the ignition condition of each spark plug appears in the sample signal. After removing outliers, we achieved $705$ samples from all engine conditions. One typical sample signal is depicted in Figure \ref{fig4.views}. All the analyses explained in the following sections were conducted on the sample signals.

\section{Proposed Method}   \label{section.pm}
\paragraph{}
Our research presents a multi-modal Autoencoder with a novel training procedure for multi-modal data fusion for intelligent condition monitoring and fault diagnosis that has yet to be proposed in previous research. Our objective is to take data fusion to the next level by enabling the data fusion method to not only combine two modalities but also allow one of these modalities to be omitted intentionally or accidentally during the test time with a very slight performance reduction or none at all. We aim to learn a common representation from two modalities describing an event such that: (1) any single modality can be predicted from the common representation, (2) any missing single modality can be predicted from the common representation consisting only of the other present modality and (3) like CCA, the learned modality-specific representations to be maximally similar. The first and second goals can be achieved together by a multi-modal Autoencoder, but projecting two modalities into a common subspace is not guaranteed. To this end, we propose a novel objective function for training the multi-modal Autoencoder, which can achieve all three above-mentioned goals. By satisfying the above-mentioned constraints, an enriched common representation containing the information of all the input modalities is achieved. After representation learning by Autoencoder, the decoders are omitted, and the latent space is used for equipment health state classification downstream task. 
\paragraph{}
In this study, vibration and acoustic signals are regarded as two modalities for engine condition monitoring and spark plug fault diagnosis. The explanation of the dataset is available in section \ref{section.es}. Our presented model and the corresponding training procedure are explained in the following sub-sections.

\subsection{Model Architecture}       \label{section.pa.arch}
\begin{figure}[t!]
	\centering
	\includegraphics[height=0.5\textwidth]{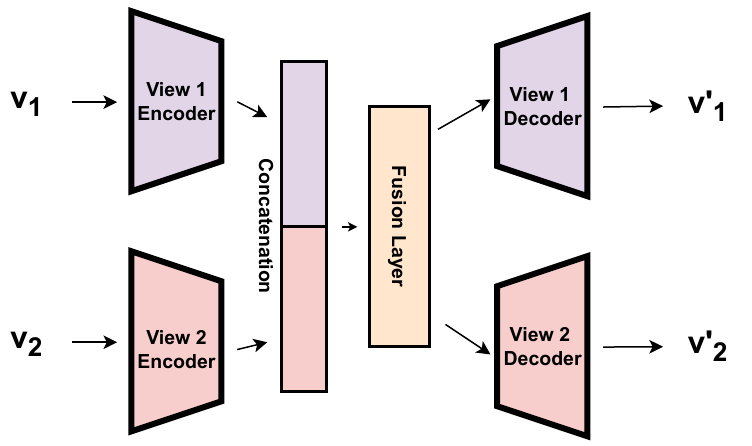}\caption{Multi-Modal Autoncoder architecture}\label{fig5}
	\centering
\end{figure}

\paragraph{}
The abstract presented multi-modal Autoencoder architecture is depicted in Figure \ref{fig5}. In the presented architecture, which complies with the one-view-one-net strategy \cite{16}, modality-specific encoders are dedicated to extract features from the corresponding modality. Modality-specific extracted features are then fused by a fusion layer consisting of several fully connected layers followed by a nonlinear activation function (such as ReLU) and form a common representation from two input modalities. Afterward, each modality-specific decoder is responsible for reconstructing the corresponding modality from the common representation.
\paragraph{}
More formally, consider a multi-modal sample of the dataset $z = (v_1, v_2)$; Different input modalities may have different data types, such as text and image, and the input vectors to modality-specific encoders may have different sizes. $(v_1, 0)$ and $(0, v_2)$ denote the inputs consisting only of the information of one modality while the other modality is omitted (a constant zero vector is passed to the omitted modality encoder as input). For simplicity, we name the modality $1$ encoder as $E_1(.)$, modality $2$ encoder as $E_2(.)$, modality $1$ decoder as $D_1(.)$, modality $2$ decoder as $D_2(.)$, and the fusion layer as $FL(.)$, which $E_1, E_2, D_1, and D_2$ are neural networks consisting of appropriate neural layers according to the input data type and $FL$ is a neural network consisting of fully connected layers. Given the multi-modal input $z = (v_1, v_2)$, the encoders and fusion layer compute the common representation as follows:
\begin{equation}\label{eq:1}
	h(z) = FL((E_1(v_1), E_2(v_2)))
\end{equation}
We call the output of the fuse layer ($FL$) the common representation, latent representation, and code layer exchangeably.

\begin{figure}[h!]
	\centering
	\subfigure[Joint-modal input]{\includegraphics[width=0.3\textwidth]{IMAGES/6-generaljoint.pdf}}
	\hfill
	\subfigure[Single-modal $v_1$ input]{\includegraphics[width=0.3\textwidth]{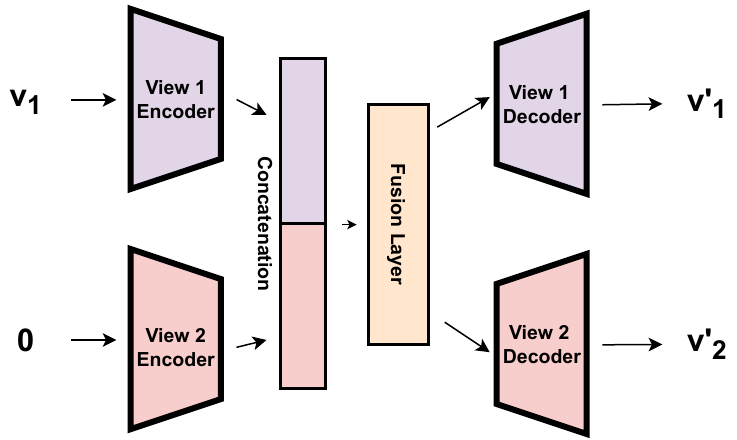}}
	\hfill
	\subfigure[Single-modal $v_2$ input]{\includegraphics[width=0.3\textwidth]{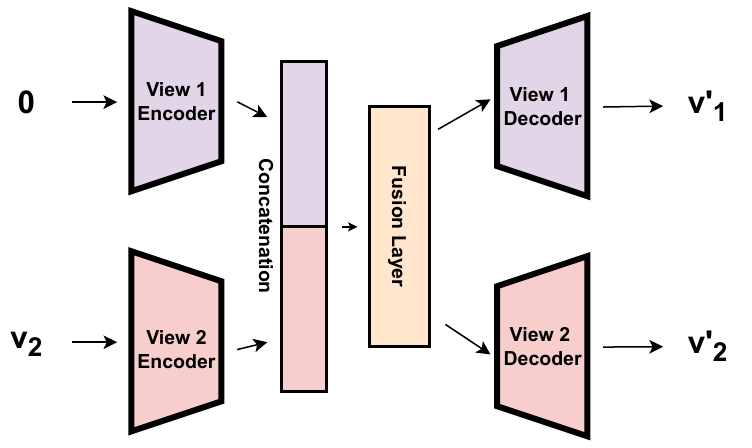}}
	\caption{3 modes of passing the inputs to network}\label{fig6.modes}
\end{figure}

The decoders then are supposed to reconstruct the input $z$ from the extracted common representation $h(z)$ by computing:
\begin{equation}\label{eq:2}
	z' = D(h(z)) = (D_1(h(z)), D_2(h(z)))
\end{equation}
where $z'$ is reconstruction of $z$.
\paragraph{}
The multi-modal samples are input to the network in 3 modes:
\begin{itemize}
	\item joint-modal $(v_1, v_2)$
	\item single-modal $v_1$ $(v_1, 0)$
	\item single-modal $v_2$ $(0, v_2)$
\end{itemize}
In the joint-modal mode, both of the input modalities are present. In any single-modal mode, only one modality is present and is passed to the network, and the other one is omitted and a constant zero vector is passed to the missing modality encoder. We call the single-modal input passing mode the missing modality case as well, especially during the test time, in which one of the input modalities is missed. A common representation is achieved from each of the three above-mentioned input passing modes for each sample. The three modes of passing input to network architecture are depicted in Figure \ref{fig6.modes}. We outline the training procedure for training the presented architecture in the following sub-section.

\subsection{Training Strategy}         \label{section.pa.loss}
\paragraph{}
More formally restating our goals, given a multi-modal dataset $Z={\{z_i\}_{i=1}^N}={\{(v_1^i, v_2^i)\}_{i=1}^N}$ consisting of $N$ samples with two $v_1$ and $v_2$ modalities, for each sample, $(v_1^i, v_2^i)$, we would like to:
\begin{enumerate}
	\item  Minimize the self-reconstruction error, i.e., minimize the error in reconstructing $v_1^i$ from $v_1^i$ and $v_2^i$ from $v_2^i$ in the joint-modal input passing.
	
	\item  Minimize the cross-reconstruction error, i.e., minimize the error in reconstructing $v_1^i$ from $v_2^i$ and $v_2^i$ from $v_1^i$ in the single-modal input passing.
	
	\item  Minimize the Euclidean distance between the single-modal input $v_1^i$ latent representation ($h(v_1^i)$) and the single-modal $v_1^j$ latent representations ($h(v_1^j)$) of all other samples in the dataset (except itself) if $i$ and $j$ belong to the same class and maximize the distance if $i$ and $j$ belong to different classes.
	
	\item  Minimize the distance between the single-modal input $v_2^i$ latent representation ($h(v_2^i)$) and the single-modal $v_2^j$ latent representations ($h(v_2^j)$) of all other samples in the dataset (except itself) if $i$ and $j$ belong to the same class and maximize the distance if $i$ and $j$ belong to different classes.
	
	\item  Minimize the distance between the single-modal input $v_1^i$ latent representation ($h(v_1^i)$) and the single-modal $v_2^j$ latent representations ($h(v_2^j)$) of all other samples in the dataset if $i$ and $j$ belong to the same class and maximize the distance if $i$ and $j$ belong to different classes.
	
	\item  Minimize the distance between the joint-modal representations $h(z^i)$ and $h(z^j)$ of inputs $z^i$ and $z^j$ ($i\neq j$) if $i$ and $j$ belong to the same class and maximize the distance if $i$ and $j$ belong to different classes.	
\end{enumerate}
\paragraph{}
We use $h(v_1^i)$ and $h(v_2^i)$ to denote representations $h((v_1^i), 0)$ and $h((0, v_2^i))$ for brevity and simplicity. For a given multi-modal sample with $v_1$ and $v_2$ modalities, $h(v_1)$ means that we are computing the common latent representation using only the single modality $v_1$; In other words, we set $v_2=0$ in Equation \ref{eq:1} and a constant zero vector is passed to the corresponding encoder.
\paragraph{}
The two first goals guarantee the predictability of one modality from the same and the other modality for each sample. The second three goals state that for a typical sample $i$, not only $h(v_1^i)$ and $h(v_2^i)$ should have a small distance (be similar), but also $h(v_1^i)$ should have a small distance with $h(v_1^j)$ and $h(v_2^j)$ of all other samples $j$ in the data set if $i$ and $j$ belong to the same class and $h(v_1^i)$ should have large distance with $h(v_1^j)$ and $h(v_2^j)$ if $i$ and $j$ belong to different classes. The same statement is held for $h(v_2^i)$ and $h(v_1^j)$, $h(v_2^j)$. The last goal states that the joint-modal representation of each sample $i$ should be attracted to the joint-modal representation of other samples belonging to the same class and should be repelled by the joint-modal representation of samples belonging to a class rather than the one $i$ belongs to. In other words, the last four goals force all the three joint-modal, single-modal $v_1$, and single-modal $v_2$ representations of instances belonging to the same category to be more similar and force the representations of instances belonging to different categories to be dissimilar.
\paragraph{}
To realize the last four goals, we utilize a contrastive loss inspired by the one presented in \cite{55}. We aggregate all the above-mentioned goals into a single objective function and achieve these goals by finding the presented multi-modal Autoencoder architecture parameters $\theta$, which minimizes Equation \ref{eq:8}, consisting of different terms described in Equations \ref{eq:3}-\ref{eq:7}.
\begin{equation}\label{eq:3}
	\Im_1(\theta) = \sum_{i=1}^{N}L(z^i,D(h(\tilde{z}^i))+\delta_1L(z^i,D(h(\tilde{v}_1^i))+\delta_2L(z^i,D(h(\tilde{v}_2^i))
\end{equation}
\begin{equation}\label{eq:6}
	\tilde{z}=(\tilde{v}_1,\tilde{v}_2) \leftarrow (v_1,v_2)+(n_{v_1}, n_{v_2})
\end{equation}
Where $L$ is a reconstruction loss (Mean Square Error (MSE) is used), $N$ denotes the number of samples within the mini-batch, ${\delta}_1$ and ${\delta}_2$ are the scaling factors to balance the second and the third terms in Equation \ref{eq:3} with respect to the remaining first term. Equation \ref{eq:6} states that $v_1$ and $v_2$ modalities are corrupted by noise vectors randomly sampled from a uniform distribution within the range of $(-0.05, 0.05)$ with corresponding dimensions before being passed to the Autoencoder. Note that Equation \ref{eq:3} is the same loss function as the one used in \cite{22, 24} for unsupervised representation learning. The first term in Equation \ref{eq:3} is the loss function of the vanilla multi-modal Autoencoder without taking missing modality considerations into account, which is only supposed to learn meaningful representation from the multiple input modalities and reconstruct the inputs from the extracted representation with low reconstruction error. The second two terms in Equation \ref{eq:3} ensure that in case of having missing modality, both modalities can be predicted from the common representation of the present modality. 
\paragraph{}
Since we aim to solve a classification task, we added two supervised contrastive terms as described in Equations \ref{eq:4}-\ref{eq:7}.
\begin{equation}\label{eq:4}
	\Im_2(\theta) = \sum_{i=1}^{N} \sum_{j=1}^{N}I({c_i}, {c_j})\|h(\tilde{z}_i^{c_i}), h(\tilde{z}_j^{c_j})\|_2
\end{equation}
\begin{equation}\label{eq:5}
	\Im_3(\theta) = \sum_{m=1}^2 \sum_{n=1}^2 \sum_{i=1}^{N} \sum_{j=1}^{N}I({c_i}, {c_j})\|h(\tilde{v}_m^{(i, c_i)}), h(\tilde{v}_n^{(j, c_j)})\|_2
\end{equation}
\begin{equation}\label{eq:7}
	I({c_i}, {c_j})=
	\begin{cases}
		-1 & \text{if } c_i \neq c_j\\
		1 & \text{if } c_i = c_j
	\end{cases}
\end{equation}
Where $c_i$ denotes the class label of sample $i$. In Equation \ref{eq:5}, if $m=n$ then $i \neq j$. Note that Equations \ref{eq:4} and \ref{eq:5} are calculated in vectorized fashion and are not computationally expensive. These terms are the supervised contrastive terms that are responsible for realizing goals $3-6$ and force the three $h(z), h(v_1)$, and $h(v_2)$ representations of samples belonging to the same class to form a dense cluster in the latent space far from representation cluster of other classes. 
\paragraph{}
Equation \ref{eq:8} describes the loss function.
\begin{equation}\label{eq:8}
	\Im(\theta) =  \Im_1(\theta) + {\lambda}_1 \Im_2(\theta) + {\lambda}_2 \Im_3(\theta))
\end{equation}
Where ${\lambda}_1$ and ${\lambda}_2$ are the scaling factors to balance the second and the third terms in the objective function with respect to the remaining first term. $h(z)$, $h(v_1)$, and $h(v_2)$ are not guaranteed to be identical or maximally similar, but the training procedure due to the existence of various reconstructive and contrastive terms in the loss function makes them that way. In conclusion, Equation \ref{eq:3} is the loss function of a vanilla multi-modal Autoencoder while taking missing modality considerations into account. Equation \ref{eq:4} forces the joint-modal representation of all samples belonging to the same class to be more compact, which leads to better data fusion and more accurate classification performance in the presence of both of the input modalities. Equation \ref{eq:5} forces the different single-modal representations ($h(v_1)$ and $h(v_2)$) of all the samples belonging to the same class to be maximally aligned and have small distance, which means not only the two single-modal representations of the same sample are maximally aligned, but also the two single-modal representations of all samples belonging to the same class are more compact and form a dense cluster far from the representation of samples belonging to other classes, which leads to more robustness and better classification performance. The pseudo-codes for training the multi-modal Autoencoder, classifier, and inference are presented in Algorithms \ref{alg1}-\ref{alg3}.

\begin{algorithm}
	\caption{Mini-batch stochastic gradient descent training of multi-modal Autoencoder}\label{alg1}
	\begin{algorithmic}
		
		\For{number of training iterations}
		\begin{itemize}
			\item Sample mini-batch of N multi-modal data $\{z^i\}_{i=1}^N=\{(v_1^i, v_2^i)\}_{i=1}^N$
			\item Get $h(z^i), h(v_1^i), h(v_2^i)$  using $E_1$, $E_2$, and $FL$.
			\item Reconstruct $\hat{z}^i_{z^i}, \hat{z}^i_{v_1^i}, \hat{z}^i_{v_2^i}$ from $h(z^i), h(v_1^i), h(v_2^i)$ using $D_1$ and $D_2$.
			\item Update the multi-modal autoencoder weights by descending their stochastic gradient:
		\end{itemize} \\
		\begin{center}
			$\frac{1}{N} \nabla_\theta \Im_{\theta}$
		\end{center}
		\EndFor
	\end{algorithmic}
\end{algorithm}

\begin{algorithm}
	\caption{Mini-batch stochastic gradient descent training of latent representation classifier}\label{alg2}
	\begin{algorithmic}
		\item Choose your input passing mode among one of the following items:
		\item $z: (v_1, v_2)$, $v_1: (v_1, 0)$, and $v_2: (0, v_2)$
		\For{all the samples in the train set}
		\begin{itemize}
			\item Get the latent representation of the input sample using $E_1$, $E_2$, and $FL$, while the parameters of these networks are kept frozen.
		\end{itemize}
		\EndFor
		
		\For{number of classifier training iterations}
		\begin{itemize}
			\item Train a single fully connected layer with linear activation function on the acquired representations of the train set samples.
		\end{itemize}
		\EndFor
	\end{algorithmic}
\end{algorithm}

\begin{algorithm}[!h]
	\caption{Inference procedure}\label{alg3}
	\begin{algorithmic}
		\item Choose your input passing mode among one of the following items:
		\item $z: (v_1, v_2)$, $v_1: (v_1, 0)$, and $v_2: (0, v_2)$
		\item Utilize the classifier \emph{C} which is previously trained on one of the $h(z)$, $h(v_1)$, or $h(v_2)$ representations.
		\For{Each query sample in the test set}
		\begin{itemize}
			\item Get the latent representation of the input sample using $E_1$, $E_2$, and $FL$, while the parameters of these networks are kept frozen.
			\item predict the class label for the input sample using the acquired latent representation of the sample and the classifier \emph{C}.
		\end{itemize}
		\EndFor
	\end{algorithmic}
\end{algorithm}
\paragraph{}
The loss function in general enables us to fuse two modalities of data and maximizes the predictability of the missing modality from the representation of the present one. The missing modality case presents a situation in which one of the modalities is omitted deliberately in order to reduce production costs, or one of the modalities is omitted due to sensor failure occurrence. As a merit of our unique training procedure, the representation of the present modality is rich enough for precise condition monitoring in missing modality case, and the multi-modal condition monitoring system will perform more robustly. Also, one can deliberately omit the modality coming from the more expensive sensor or the one with more complex installation in test time (in the production line) but achieve similar performance as using both. As a result of this technique, we take the data fusion in the machine health monitoring realm to a higher level, which means not only we are fusing different input data modalities to achieve better diagnosis performance than using a single-modal input but also one of these input modalities can be omitted in test time whether deliberately or inadvertently and still achieve similar performance as using both with slight sacrifice in performance or none at all.
\paragraph{}
In conclusion, the proposed training strategy enables the multi-modal intelligent condition monitoring systems to use more information during the test time without imposing additional costs of real sensors in the production line. We use ADAM optimizer to find optimal parameters. The model has $5$ hyper-parameters: $(1)$ the number of trainable parameters in each modality encoder or decoder or fuse layer, $(2)$ and $(3)$ $\lambda_1$ and $\lambda_2$, $(4)$ and $(5)$ $\delta_1$ and $\delta_2$, $(6)$ batch size, and $(7)$ the optimizer learning rate. The first one depends on the specific task at hand and can be tuned using a validation set. The second and third hyper-parameters ensure that the reconstruction error terms have the same range as similarity terms, which are easy to approximate using the given data. The fourth and fifth hyper-parameters balance different reconstructive error terms in Equation \ref{eq:3}, which are set equal to $1$ in current research. The last two hyper-parameters are common for all deep learning based solutions.
\section{Results and Discussion}   \label{section.randd}
\subsection{Simulation Setup}  \label{section.randd.ss} 
\begin{figure*}[h!]
	\centering
	\includegraphics[height=0.4\textwidth]{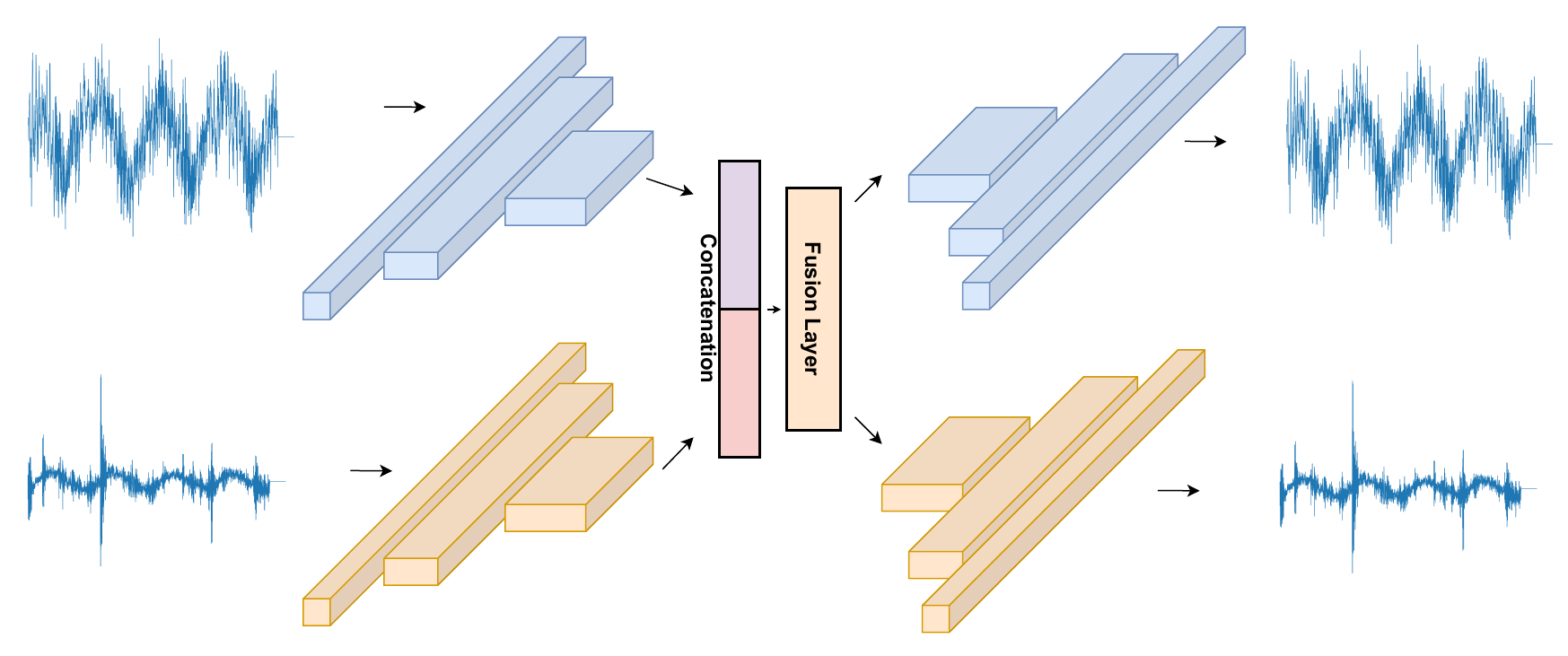}\caption{Multi-Modal Autoncoder architecture with CNN layers}\label{fig85}
	\centering
\end{figure*}

\paragraph{}				
Based on the abstract architecture presented in section \ref{section.pa.arch}, we designed an architecture consisting of one-dimensional convolutional and one-dimensional transpose convolutional layers for encoders and decoders, respectively, for representation learning and evaluating the presented methodology on the spark plug fault diagnosis case study. The designed architecture is depicted in Figure \ref{fig85}.
\paragraph{}
Acoustic and vibration signals are regarded as two input modalities that we aim to fuse and learn common representation from. For the rest of this article, we use "a" and "v" to denote acoustic and vibration modalities, respectively. Since we use a dataset with extremely limited samples, we randomly choose $40$ samples for validation set at first, $10$ from each class, and we use the remaining set for training. We find the network optimal hyper-parameters, such as the number of layers, number of kernels, kernel size, etc., based on the network's performance on the validation set. Afterward, for training and evaluating the presented methodology, we use $7$-fold stratified cross-validation on the remaining set since we are dealing with a dataset with very limited samples, and we still use the validation set for early stopping on each fold. In addition to the early stopping, noise projection and weight decay are used as regularization terms. The architecture of encoders, decoders, and the fuse layer are presented in Tables \ref{Tableencarch}-\ref{Tablefusearch}.
\begin{table*}[h!]\centering
	\normalsize
	\caption{Encoder architecture}
	\resizebox{\columnwidth}{!}{%
		\begin{tabular}{|p{5cm}|c|c|c|c|c|c|c|c|c|}
			\hline
			& Layer 1 & Layer 2 & Layer 3 & Layer 4 & Layer 5 & Layer 6 & Layer 7 & Layer 8 \\
			\hline\hline
			(Convolution kernel size, stride) & (11, 1) & (6, 1) & (6, 1) & (6, 1) & (6, 1) & (6, 1) & (70, 1) & Flatten \\
			\hline
			Number kernels & 10 & 20 & 40 & 60 & 80 & 100 & 128 & - \\
			\hline
			(Pooling kernel size, stride) & (2, 2) & (2, 2) & (2, 2) & (2, 2) & (2, 2) & (2, 2) & - & - \\
			\hline
			Activation function & ReLU & ReLU & ReLU & ReLU & ReLU & ReLU & - & - \\
			\hline
			Input size & (-1, 1, 4800) & (-1, 10, 2395) & (-1, 20, 1195) & (-1, 40, 595) & (-1, 60, 295) & (-1, 80, 145) & (-1, 100, 70) & (-1, 128, 1)\\
			\hline
			Output size & (-1, 10, 2395) & (-1, 20, 1195) & (-1, 40, 595) & (-1, 60, 295) & (-1, 80, 145) & (-1, 100, 70) & (-1, 128, 1) & (-1, 128) \\
			\hline
		\end{tabular}%
	}
	\label{Tableencarch}
\end{table*}
\begin{table*}[h!]\centering
	\normalsize
	\caption{Decoder architecture}
	\resizebox{\columnwidth}{!}{%
		\begin{tabular}{|p{5cm}|c|c|c|c|c|c|c|c|c|}
			\hline
			& Layer 1 & Layer 2 & Layer 3 & Layer 4 & Layer 5 & Layer 6 & Layer 7 & Layer 8 \\
			\hline\hline
			(Convolution kernel size, stride) & Reshape & (70, 1) & (6, 1) & (6, 1) & (6, 1) & (6, 1) & (6, 1) & (11, 1) \\
			\hline
			Number kernels & - & 100 & 80 & 60 & 40 & 20 & 10 & 1 \\
			\hline
			(Unpooling kernel size, stride) & - & (2, 2) & (2, 2) & (2, 2) & (2, 2) & (2, 2) & (2, 2) & - \\
			\hline
			Activation function & - & ReLU & ReLU & ReLU & ReLU & ReLU & ReLU & - \\
			\hline
			Input size & (-1, 128) & (-1, 128, 1) & (-1, 100, 140) & (-1, 80, 290) & (-1, 60, 590) & (-1, 40, 1190) & (-1, 20, 2390) & (-1, 10, 4790)\\
			\hline
			Output size & (-1, 128, 1) & (-1, 100, 140) & (-1, 80, 290) & (-1, 60, 590) & (-1, 40, 1190) & (-1, 20, 2390) & (-1, 10, 4790) & (-1, 1, 4800) \\
			\hline
		\end{tabular}%
	}
	\label{Tabledecarch}
\end{table*}
\begin{table*}[h!]\centering
	\caption{Fuse Layer architecture}
	\begin{tabular}{|c|c|c|}
		\hline
		& Layer 1 & Layer 2 \\
		\hline
		(Input, Output) & (256, 128) & (128, 128) \\
		\hline
		Layer type & Dense & Dense \\
		\hline
		Activation function & ReLU & - \\
		\hline
	\end{tabular}
	\label{Tablefusearch}
\end{table*}

\subsection{Simulation Results}			\label{section.randd.sr}
\begin{figure}
	\centering
	\subfigure[Joint-modal input common representation]{\includegraphics[width=0.3\textwidth]{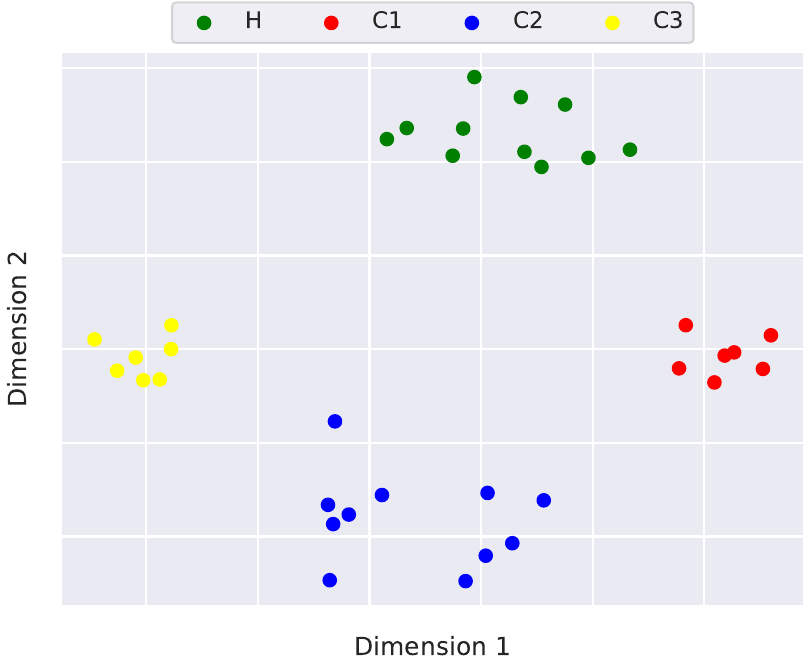}}
	\hfill
	\subfigure[Acoustic single-modal input common representation]{\includegraphics[width=0.3\textwidth]{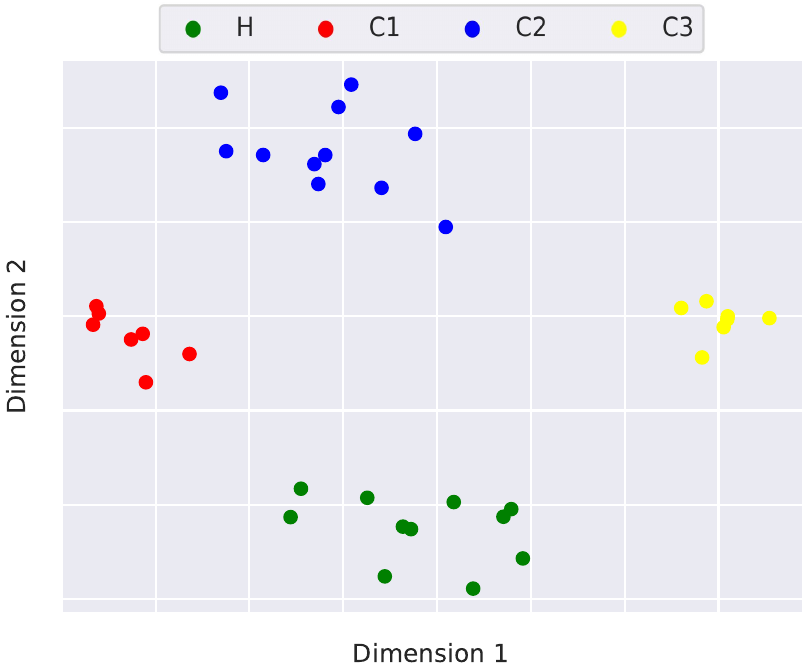}}  
	\hfill
	\subfigure[Vibration single-modal input common representation]{\includegraphics[width=0.3\textwidth]{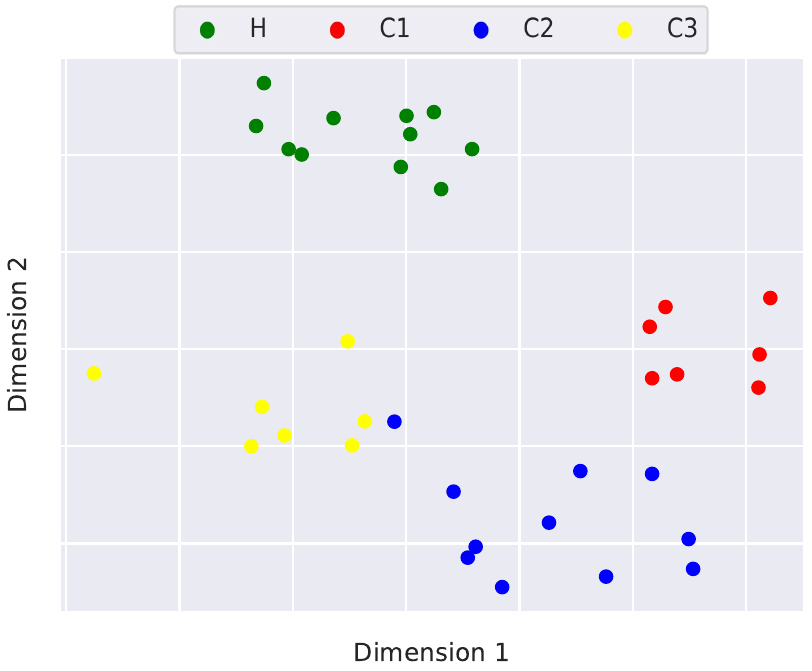}}
	\caption{Three representations of network code layer}\label{fig4.representations}
\end{figure}

\paragraph{}					
For latent space investigation, the remaining set is randomly divided into train and test splits, with $0.95$ and $0.05$ proportions, respectively, and after training the network on the train split, $2$D visualizations of the common representation of joint-modal, acoustic single-modal, and vibration single-modal inputs of the test split samples are obtained by t-sne \cite{56} and are depicted in Figure \ref{fig4.representations}. 

\begin{figure}[b!]
	\centering
	\subfigure[$h(a)$, and $h(v)$ representations]{\includegraphics[width=0.48\textwidth]{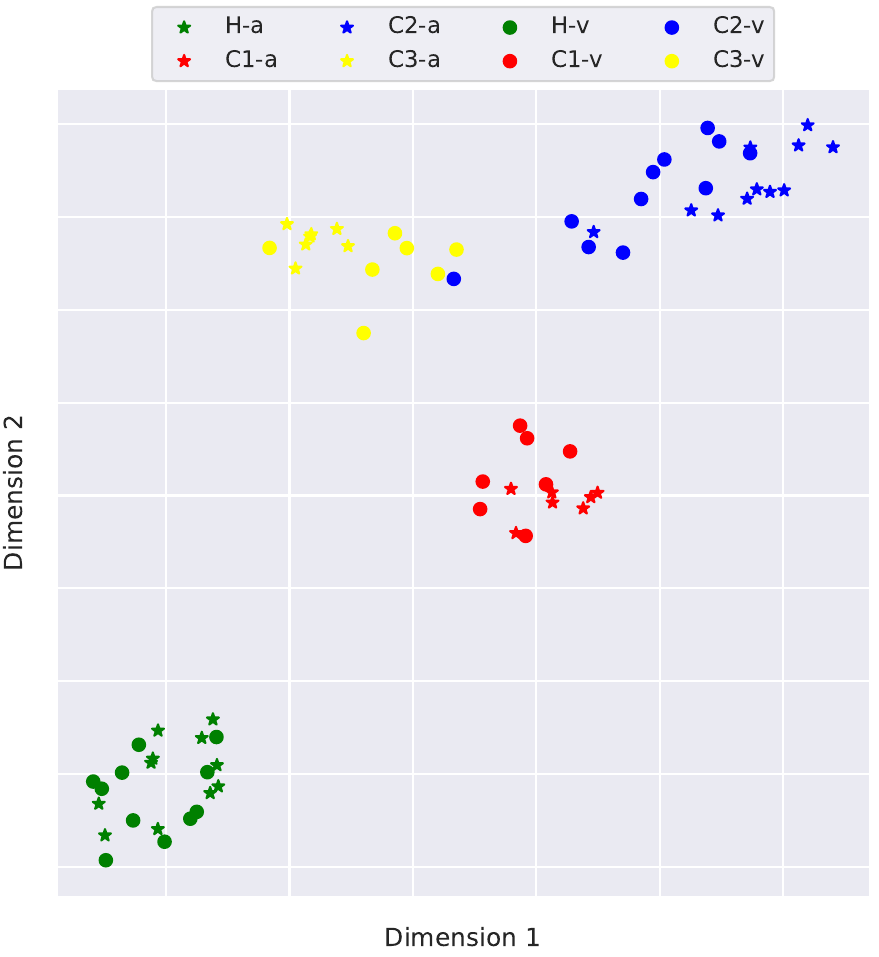}}
	\hfill
	\subfigure[$h(a)$, $h(v)$, and $h(z)$ representations]{\includegraphics[width=0.48\textwidth]{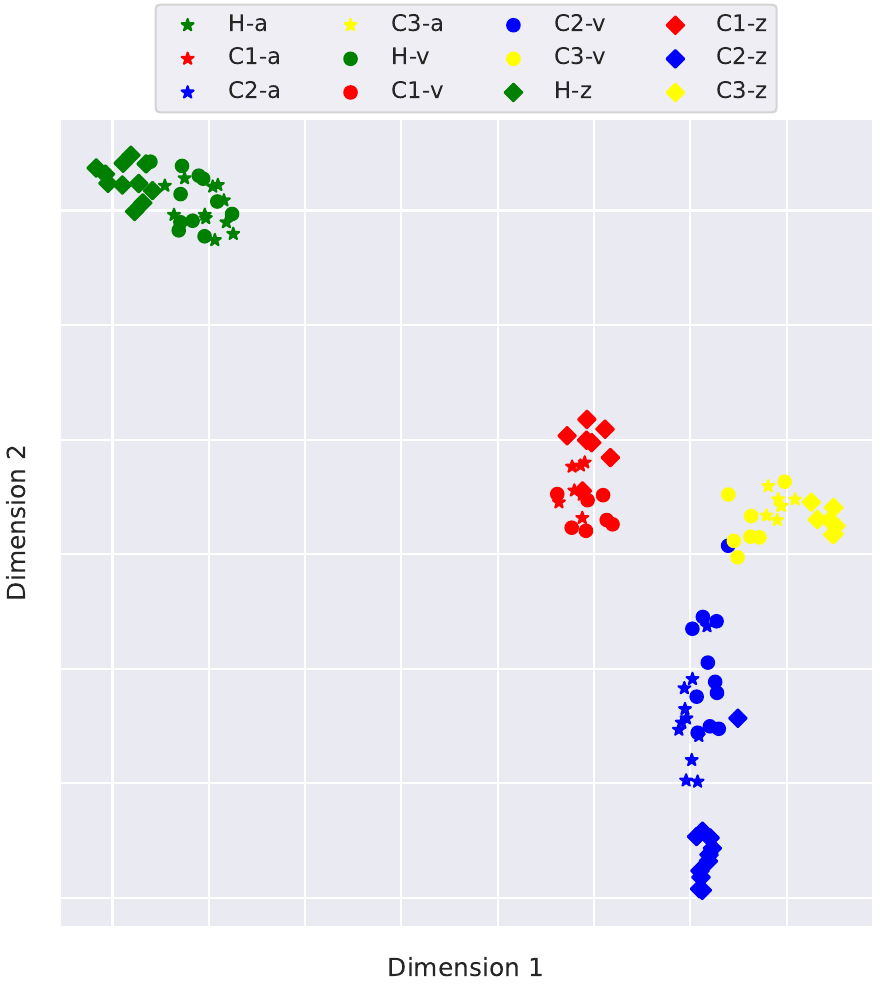}}
	\caption{Similarity of $h(a)$, $h(v)$, and $h(z)$ representations in latent space}\label{fig5.representations}
\end{figure}

\begin{figure}[h!]
	\centering
	\subfigure[Test on $h(z)$ confusion matrix]{\includegraphics[width=0.3\textwidth]{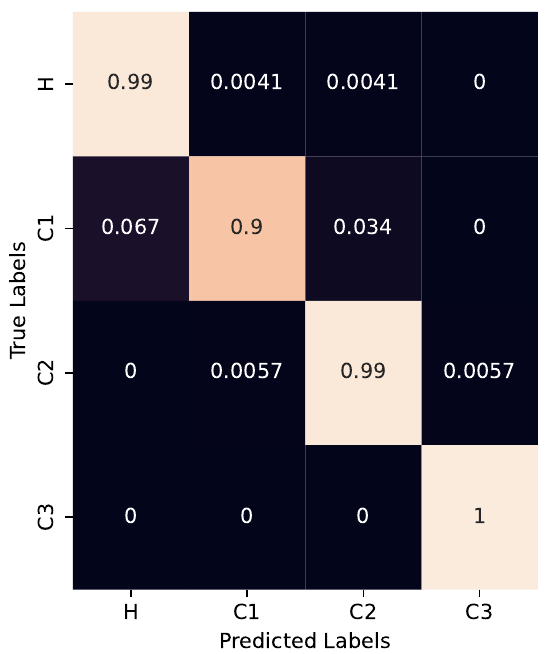}}
	\hfill
	\subfigure[Test on $h(a)$ confusion matrix]{\includegraphics[width=0.3\textwidth]{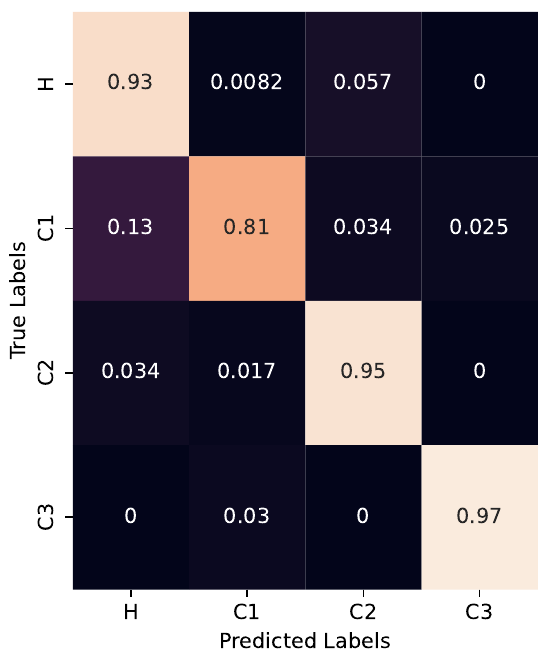}}
	\hfill
	\subfigure[Test on $h(v)$ confusion matrix]{\includegraphics[width=0.3\textwidth]{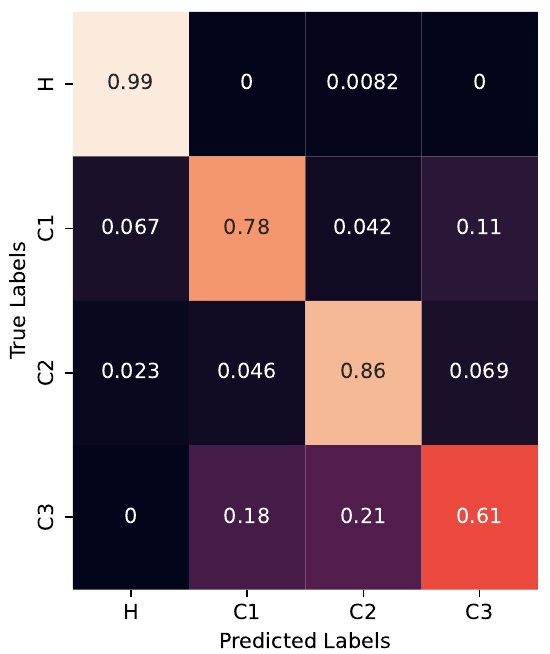}}
	\caption{Experiment 1 results}\label{fig5.cms}
\end{figure}

Moreover, a $2$D visualization of the acoustic single-modal and vibration single-modal latent representations in the same plot is depicted in Figure \ref{fig5.representations}-a, and a $2$D visualization of all the three common representations in the same plot is depicted in Figure \ref{fig5.representations}-b.
\paragraph{}
Figure \ref{fig4.representations}-a depicts that the goal number $6$ in section \ref{section.pa.loss} is realized, and the joint-modal representation of different samples belonging to the same class has formed a dense cluster far from clusters of the other classes. Figure \ref{fig4.representations}-b depicts that the goal number $3$ in section \ref{section.pa.loss} is realized, and the acoustic single-modal representation of different samples belonging to the same class has formed a dense cluster far from the cluster of other classes. Figure \ref{fig4.representations}-c depicts that the goal number $4$ in section \ref{section.pa.loss} is realized, and the vibration single-modal representation of different samples belonging to the same class has formed a dense cluster far from the cluster of other classes.
\paragraph{}
Moreover, Figure \ref{fig5.representations}-a denotes that the goal number $5$ in section \ref{section.pa.loss} is realized, and acoustic single-modal representation ($h(a)$) and vibration single-modal representation ($h(v)$) of the samples belonging to the same class are very close. Figure \ref{fig5.representations}-b denotes that not only the single-modal representations $h(a)$ and $h(v)$ are very similar in the code layer, but also the joint-modal representation $h(z)$ is also very similar to both of $h(a)$ and $h(v)$ representations and these three representations of the samples belonging to the same class form a dense cluster far apart from the clusters of other classes which leads to a highly generalizable classification performance.
\paragraph{}
Notably, we did not explicitly force the architecture to learn representations such that $h(z)$ is similar to $h(a)$ and $h(v)$; however, the training procedure implicitly led to such representations. In conclusion, the proposed neural architecture has learned representation such that different samples of the same class are gathered together and have formed dense clusters in the code layer far apart from the clusters of the other classes. 
\paragraph{}
In order to evaluate the capability of each of $h(z)$, $h(a)$, and $h(v)$ representations for classification downstream task, we proposed $4$ experiments:
\begin{enumerate}
	\item Training a classifier on $h(z)$ representation and testing on $h(z)$, $h(a)$ and $h(v)$.
	\item Training a classifier on $h(a)$ representation and testing on $h(z)$, $h(a)$ and $h(v)$.
	\item Training a classifier on $h(v)$ representation and testing on $h(z)$, $h(a)$ and $h(v)$.
	\item Training a classifier on \{$h(a) \cup h(v)$\} representation and testing on $h(z)$, $h(a)$ and $h(v)$.
\end{enumerate}
\pagebreak
\paragraph{}
In experiment $1$, the classifier's performance on $h(z)$ denotes the ability of the proposed methodology to fuse multiple modalities of data. The performance of the classifier on $h(a)$ and $h(v)$ denotes the capability of the joint-modal representation in classifying any of the single-modal inputs in case of missing modality (sensor failure occurrence, for example). In other words, good performance of the classifier on $h(a)$ and $h(v)$ denotes that the joint-modal representation is so enriched that one of the modalities can be omitted at inference time with a slight reduction in performance or none at all. As a result, the multi-modal fault diagnosis system developed by our proposed methodology performs more robustly in case of sensor failure occurrence. Also, one can deliberately omit one of the input modalities (usually the one comes from the more expensive sensor or the one with more complex installation) in order to achieve a cost-effective or an easy-to-use fault diagnosis system for production line.
\paragraph{}
In experiments $2$ and $3$, the good cross-modality performance of the classifier (classification performance on the missing modality) clarifies that the two single-modal representations are very similar in the code layer and have a small distance. Moreover, besides the classifier performance on the missing modality, good classification performance on the joint-modal representation $h(z)$ denotes that each single-modal representation has almost the entire information of both modalities.
\paragraph{}
Experiment $4$ investigates whether the union of single-modal representations can improve the classification performance or not. In the $7$-fold cross-validation procedure, after training the multi-modal Autoencoder on the train split of each fold, we carry out the experiments $1-4$ and test the performance of the trained classifier on the representations $h(z)$, $h(a)$, and $h(v)$ of the test split samples. At the end of the stratified $7$-fold cross-validation procedure, we report the average confusion matrices of the classifier on the $7$ test splits besides the average accuracy, precision, recall, and f$1$-score performance metrics of the classifier on the $7$ test splits.
\begin{itemize}
	\item \textbf{Experiment 1 Results Investigation}
\end{itemize}
\paragraph{}	
The average confusion matrices on $h(z)$, $h(a)$, and $h(v)$ representations are depicted in Figure \ref{fig5.cms}.

\begin{figure}[h!]
	\centering
	\subfigure[Test on $h(z)$ confusion matrix]{\includegraphics[width=0.3\textwidth]{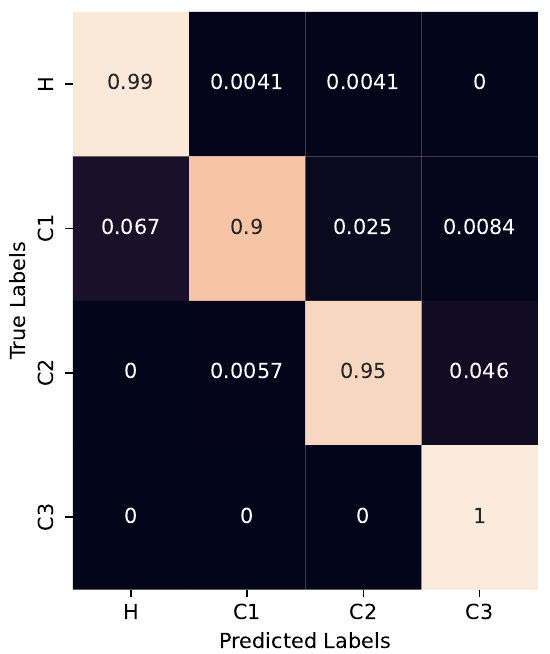}}
	\hfill
	\subfigure[Test on $h(a)$ confusion matrix]{\includegraphics[width=0.3\textwidth]{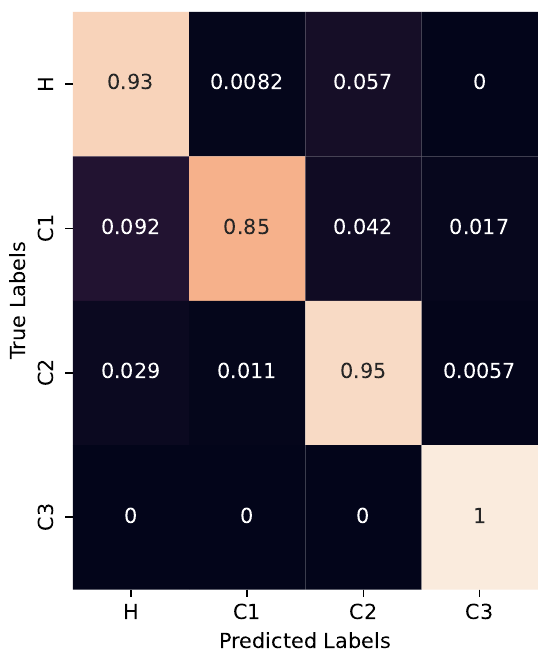}}
	\hfill
	\subfigure[Test on $h(v)$ confusion matrix]{\includegraphics[width=0.3\textwidth]{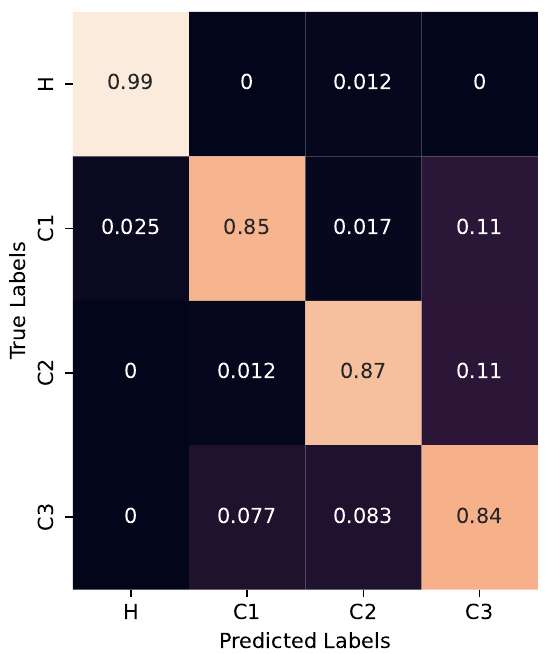}}
	\caption{Experiment 2 results}\label{fig6.cms}
\end{figure}

Figure \ref{fig5.cms}-a depicts the average confusion matrix of classification of the $h(z)$ representation of the test splits during the $7$-fold cross-validation procedure. Apparently, the performance is outstanding, which approves the ability of the proposed methodology to fuse multiple modalities into a common representation. Moreover, Figure \ref{fig5.cms}-b denotes that the knowledge of the classifier trained on $h(z)$ can be transferred to the classification task on $h(a)$ with a remarkably slight performance reduction. Although the classifier can not perfectly classify spark plug health state classes based on single-modal $h(v)$ representation, but it still achieves remarkable performance, which can be seen in Figure \ref{fig5.cms}-c. In conclusion, Figure \ref{fig5.cms} depicts that the classifier trained on joint-modal common representation has excellent capability to discriminate between different classes using both modalities or only one of them, which means not only a good fusion of different modalities has occurred but also one of the input modalities can be omitted during test time with a slight sacrifice in performance. Generally, the presence of different modalities is only necessary during the training procedure, and one of the input modalities can be omitted accidentally or deliberately during test time with a slight performance reduction or none at all.
\pagebreak
\begin{itemize}
	\item \textbf{Experiment 2 Results Investigation}
\end{itemize}
\paragraph{}
The average confusion matrices on $h(z)$, $h(a)$, and $h(v)$ representations are depicted in Figure \ref{fig6.cms}.

Figure \ref{fig6.cms}-c depicts that the classifier trained on acoustic single-modal representation $h(a)$ performs very well on single-modal vibration representation $h(v)$, which means the single-modal representation of acoustic and vibration modalities are very well gathered in the code layer. Besides, the good performance on the joint-modal representation $h(z)$ in Figure \ref{fig6.cms}-a denotes that acoustic single-modal representation has very good information from both of the modalities.

\begin{figure}[h!]
	\centering
	\subfigure[Test on $h(z)$ confusion matrix]{\includegraphics[width=0.3\textwidth]{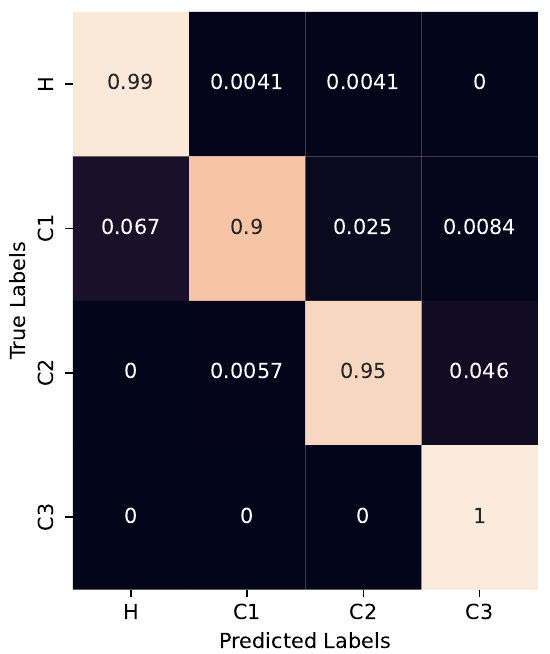}}
	\hfill
	\subfigure[Test on $h(a)$ confusion matrix]{\includegraphics[width=0.3\textwidth]{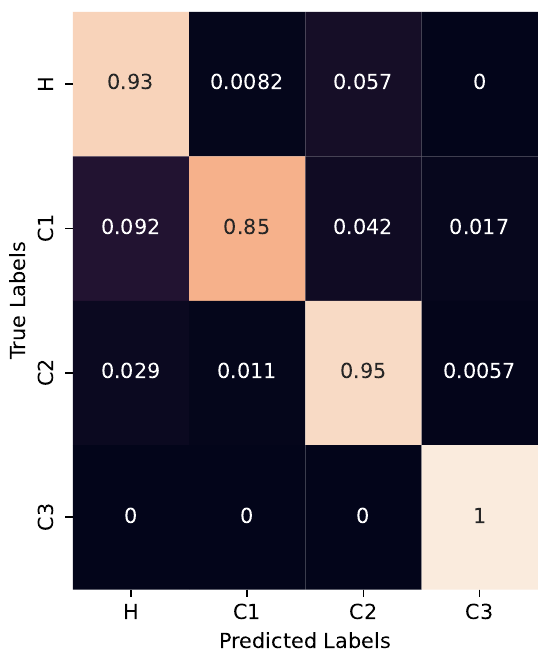}}
	\hfill
	\subfigure[Test on $h(v)$ confusion matrix]{\includegraphics[width=0.3\textwidth]{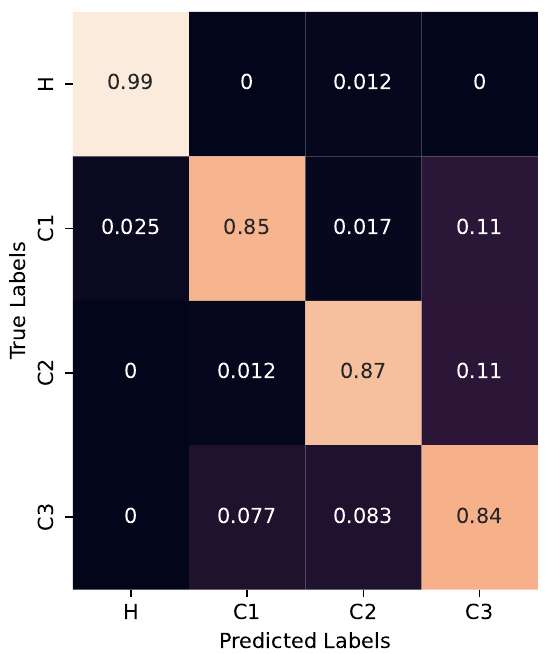}}
	\caption{Experiment 3 results}\label{fig7.cms}
\end{figure}

\begin{itemize}
	\item \textbf{Experiment 3 Results Investigation}
\end{itemize}
\paragraph{}	
The average confusion matrices on $h(z)$, $h(a)$, and $h(v)$ representations are depicted in Figure \ref{fig7.cms}.

Figure \ref{fig7.cms}-b depicts that the classifier trained on the vibration single-modal representation $h(v)$ performs very well on the single-modal acoustic representation $h(a)$, which means the single-modal acoustic and vibration representations are very well gathered in the code layer. Besides, the good performance on joint-modal representation $h(z)$ in Figure \ref{fig7.cms}-a denotes that vibration single-modal representation has very good information from both of the modalities. In conclusion, experiments $2$ and $3$ depict that $h(a)$ and $h(v)$ representations are very well gathered in the latent space, and any of these representations has a great amount of information about both since the classifier trained on the representation of each of these modalities performs well on the representation of the other.

\begin{figure}[h!]
	\centering
	\subfigure[Test on $h(z)$ confusion matrix]{\includegraphics[width=0.3\textwidth]{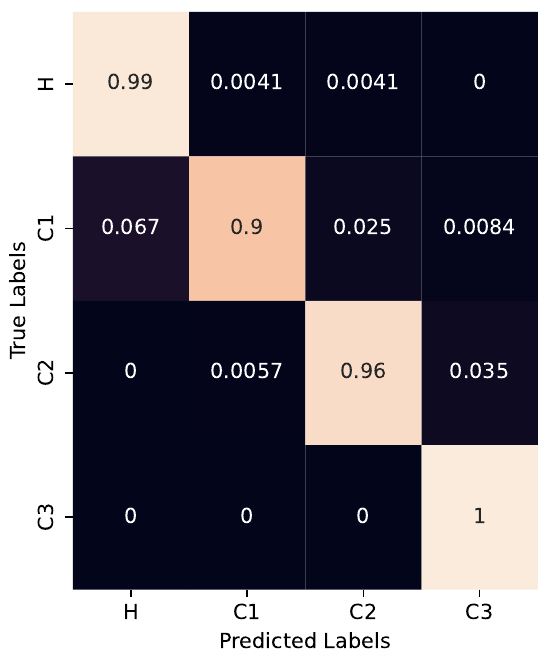}}
	\hfill
	\subfigure[Test on $h(a)$ confusion matrix]{\includegraphics[width=0.3\textwidth]{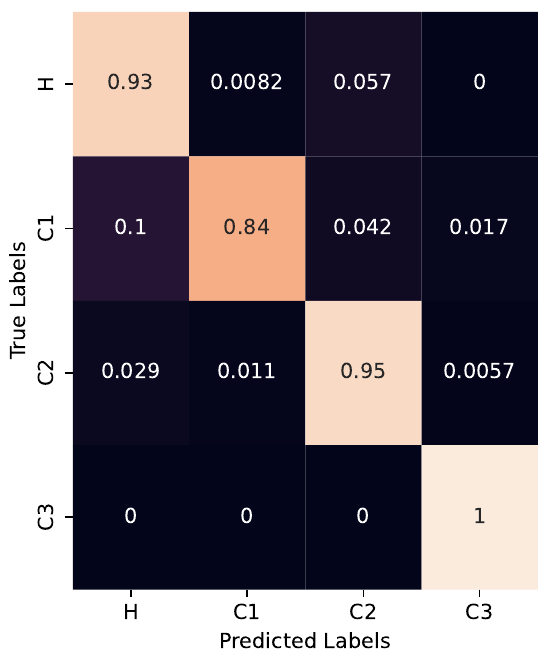}}
	\hfill
	\subfigure[Test on $h(v)$ confusion matrix]{\includegraphics[width=0.3\textwidth]{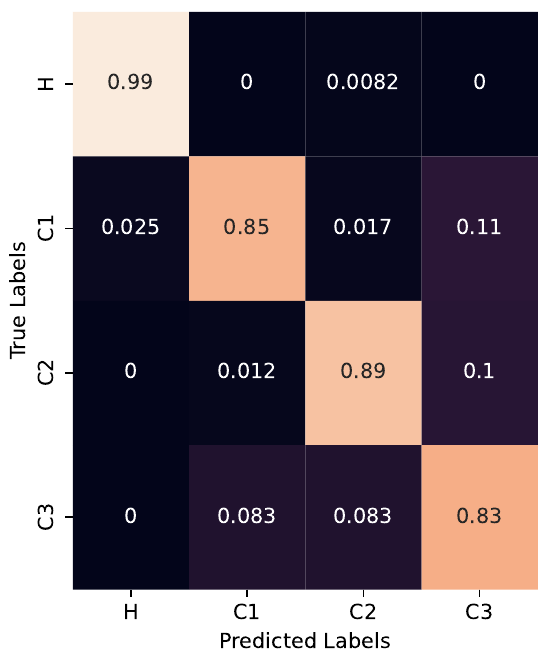}}
	\caption{Experiment 4 results}\label{fig8.cms}
\end{figure}

\begin{itemize}
	\item \textbf{Experiment 4 Results Investigation}
\end{itemize}
\paragraph{}
Experiment $4$ investigates the possibility of using the union of both of the single-modal representations in order to improve the classification performance on each of $h(z)$, $h(a)$, and $h(v)$ representations. The average confusion matrices on $h(z)$, $h(a)$, and $h(v)$ are depicted in Figure \ref{fig8.cms}.

The classification performance in experiment $4$ is almost the same as experiments $2$ and $3$.
\paragraph{}
The average performance metrics for experiments $1-4$ are evaluated based on the $7$-fold cross-validation procedure and are reported in Table \ref{Table1}.
\begin{table*}[!h]\centering
	\begin{center}
		\caption{Average performance metrics of $7$-fold cross validation procedure}
		\small\addtolength{\tabcolsep}{-4pt}
		\fontsize{7.5}{9}\selectfont
		\begin{tabular}{lccccccccccccccccccccc}
			\toprule
			Experiment  &
			\multicolumn{4}{c}{Accuracy (\%)}
			&&    \multicolumn{4}{c}{Precision (\%)}
			&&   \multicolumn{4}{c}{Recall (\%)}
			&&   \multicolumn{4}{c}{$F_1$ score (\%)}
			\\ \cline{2-5} \cline{7-10} \cline{12-15} \cline{17-19}
			
			&\multicolumn{1}{c}{$h(z)$}
			&   \multicolumn{1}{c}{$h(a)$}
			&\multicolumn{1}{c}{$h(v)$}
			
			&&&\multicolumn{1}{c}{$h(z)$}
			&   \multicolumn{1}{c}{$h(a)$}
			&\multicolumn{1}{c}{$h(v)$}
			
			&&&\multicolumn{1}{c}{$h(z)$}
			&   \multicolumn{1}{c}{$h(a)$}
			&\multicolumn{1}{c}{$h(v)$}
			
			&&&\multicolumn{1}{c}{$h(z)$}
			&   \multicolumn{1}{c}{$h(a)$}
			&\multicolumn{1}{c}{$h(v)$}
			
			&& \\  \hline
			Experiment 1 & 96.99 & 91.53 & 81.24 &&& 98.12 & 94.26 & 85.72 &&& 96.99 & 91.53 & 81.24 &&& 97.21 & 90.71 & 79.77 \\
			Experiment 2 & 95.97 & 93.44 & 88.73 &&& 97.31 & 95.62 & 89.78 &&& 95.97 & 93.44 & 88.73 &&& 96.18 & 92.52 & 88.71 \\
			Experiment 3 & 96.56 & 93.23 & 89.27 &&& 97.75 & 95.52 & 90.26 &&& 96.56 & 93.23 & 89.27 &&& 96.77 & 92.36 & 89.28 \\
			Experiment 4 & 96.27 & 93.23 & 88.98 &&& 97.56 & 95.52 & 89.97 &&& 96.27 & 93.23 & 88.98 &&& 96.48 & 92.36 & 88.95 \\
			\bottomrule
			\label{Table1}
		\end{tabular}
	\end{center}
\end{table*}

\begin{figure}[p]
	\centering
	\subfigure[Test on $h(z)$]{\includegraphics[width=0.2\textwidth]{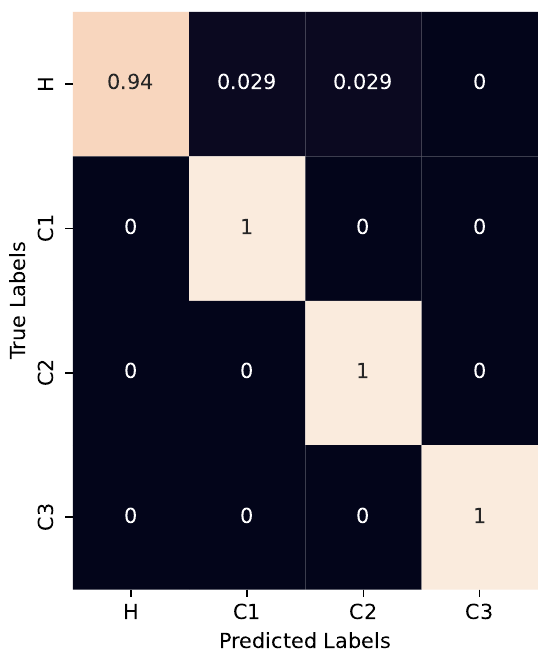}}
	\hfill
	\subfigure[Test on $h(a)$]{\includegraphics[width=0.2\textwidth]{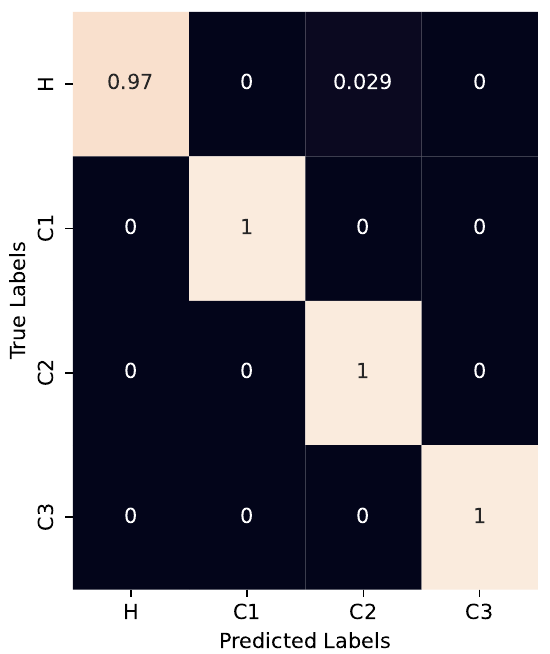}}
	\hfill
	\subfigure[Test on $h(v)$]{\includegraphics[width=0.2\textwidth]{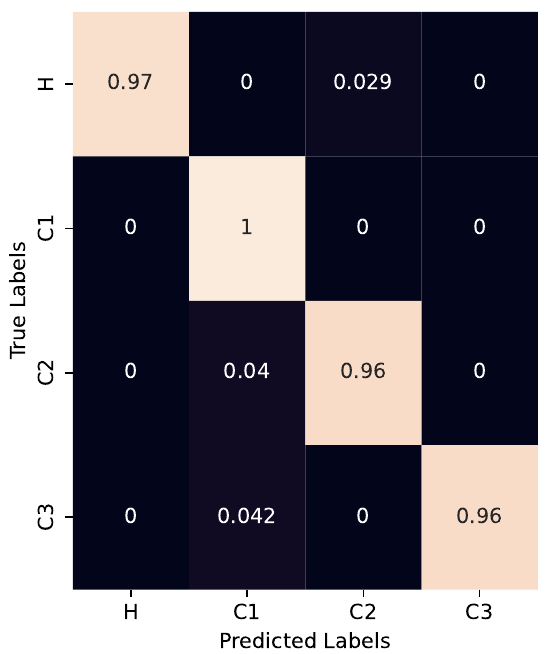}}
	\caption{Experiment 1}\label{fig9.cms}
\end{figure}

\begin{figure}[p]
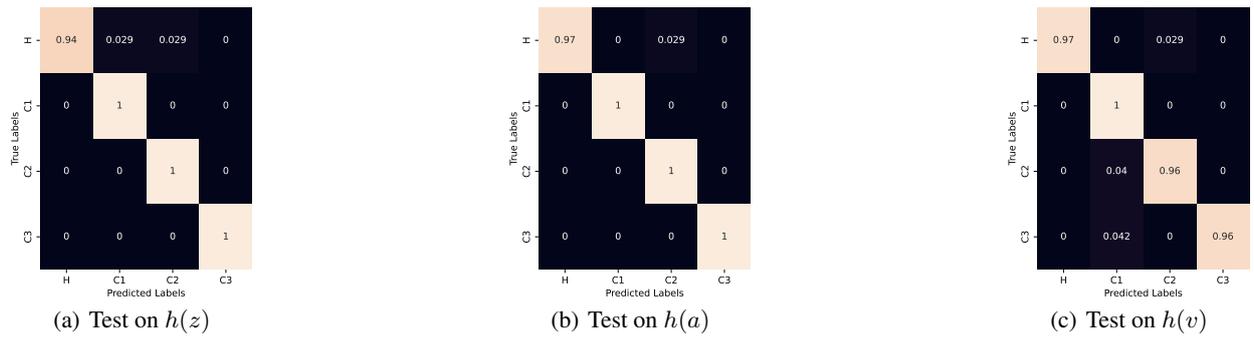

	\centering
	\subfigure[Test on $h(z)$]{\includegraphics[width=0.2\textwidth]{IMAGES/30-hv1zf.pdf}}
	\hfill
	\subfigure[Test on $h(a)$]{\includegraphics[width=0.2\textwidth]{IMAGES/31-hv1v1f.pdf}}
	\hfill
	\subfigure[Test on $h(v)$]{\includegraphics[width=0.2\textwidth]{IMAGES/32-hv1v2f.pdf}}
	\caption{Experiment 2}\label{fig10.cms}
\end{figure}

\begin{figure}[p]
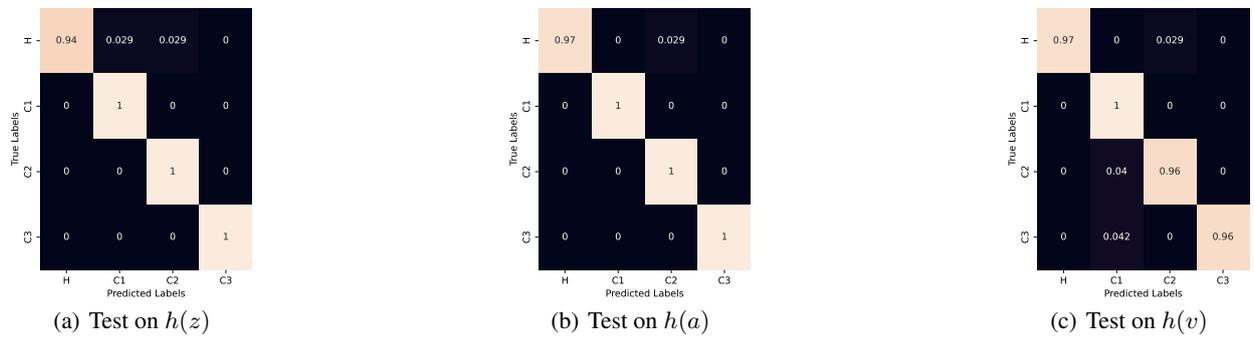

	\centering
	\subfigure[Test on $h(z)$]{\includegraphics[width=0.2\textwidth]{IMAGES/33-hv2zf.pdf}}
	\hfill
	\subfigure[Test on $h(a)$]{\includegraphics[width=0.2\textwidth]{IMAGES/34-hv2v1f.pdf}}
	\hfill
	\subfigure[Test on $h(v)$]{\includegraphics[width=0.2\textwidth]{IMAGES/35-hv2v2f.pdf}}
	\caption{Experiment 3}\label{fig11.cms}
\end{figure}

\begin{figure}[p]
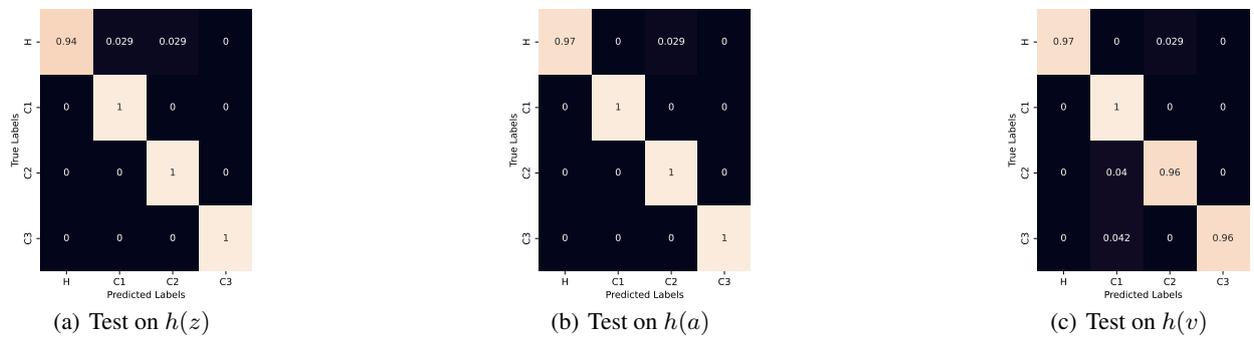

	\centering
	\subfigure[Test on $h(z)$]{\includegraphics[width=0.2\textwidth]{IMAGES/36-hv12zf.pdf}}
	\hfill
	\subfigure[Test on $h(a)$]{\includegraphics[width=0.2\textwidth]{IMAGES/37-hv12v1f.pdf}}
	\hfill
	\subfigure[Test on $h(v)$]{\includegraphics[width=0.2\textwidth]{IMAGES/38-hv12v2f.pdf}}
	\caption{Experiment 4}\label{fig12.cms}
\end{figure}

Table \ref{Table1} denotes that the classifier trained on any of $h(z)$, $h(a)$, and $h(v)$ representations performs extremely well on the other two, which means a classifier can be trained on any of $h(z)$, $h(a)$, and $h(v)$ representations during training time and be utilized to classify equipment health state based on only one of the acoustic or vibration modalities during the inference time. In other words, the latent representation acquired from any of the acoustic or vibration modalities has sufficient information about both, and one can intentionally omit one of the input modalities during the inference time in order to reduce production costs with a slight performance reduction or none at all. Moreover, the multi-modal fault diagnosis system developed based on our proposed methodology performs extremely robust in case of sensor failure occurrence. In conclusion, Table \ref{Table1} depicts that the main objective of the current research is achieved, and one of the input modalities can be omitted during the test time with a slight sacrifice in performance or none at all.
\paragraph{}
We compared the result of the current research methodology to vanilla multi-modal Autoencoder, contrastive vanilla multi-modal Autoencoder without taking missing modality considerations into account, and CorrNet \cite{24}. The average performance metrics for these methodologies are reported in Table \ref{Table2}. The neural network architecture depicted in Figure \ref{fig85} is utilized for training all of the mentioned methodologies. The objective function of the vanilla multi-modal Autoencoder is the same as Equation \ref{eq:3}. The objective function of the contrastive vanilla multi-modal Autoencoder without missing modality is depicted in Equation \ref{eq:9}.
\begin{equation}\label{eq:9}
	\Re(\theta) =  \sum_{i=1}^{N}L(z^i,D(h(z^i))) + {\alpha}_1 \Im_2(\theta)
\end{equation}
\begin{table*}[!h]\centering
	\caption{Comparison of average performance metrics based on $7$-fold cross validation evaluation performance}
	\small\addtolength{\tabcolsep}{-4pt}
	\fontsize{7.5}{9}\selectfont
	\resizebox{\columnwidth}{!}{%
		\begin{tabular}{llccccccccccccccccccccc}
			\toprule
			Methodology &
			Experiment  &
			\multicolumn{4}{c}{Accuracy (\%)}
			&&    \multicolumn{4}{c}{Precision (\%)}
			&&   \multicolumn{4}{c}{Recall (\%)}
			&&   \multicolumn{4}{c}{$F_1$ score (\%)}
			\\ \cline{3-5} \cline{7-10} \cline{12-15} \cline{17-20}
			
			&&\multicolumn{1}{c}{$h(z)$}
			&   \multicolumn{1}{c}{$h(a)$}
			&\multicolumn{1}{c}{$h(v)$}
			
			&&&\multicolumn{1}{c}{$h(z)$}
			&   \multicolumn{1}{c}{$h(a)$}
			&\multicolumn{1}{c}{$h(v)$}
			
			&&&\multicolumn{1}{c}{$h(z)$}
			&   \multicolumn{1}{c}{$h(a)$}
			&\multicolumn{1}{c}{$h(v)$}
			
			&&&\multicolumn{1}{c}{$h(z)$}
			&   \multicolumn{1}{c}{$h(a)$}
			&\multicolumn{1}{c}{$h(v)$}
			
			&& \\  \hline
			\multirow{4}{3cm}{Vanilla Multimodal AE Without missing modality}
			& Experiment 1 & 75.48 & 39.28 & 50.82 &&& 73.43 & 15.65 & 26.85 &&& 75.14 & 39.28 & 50.78 &&& 74.19 & 21.59 & 34.15 \\
			& Experiment 2 & 38.21 & 63.18 & 25.14 &&& 40.71 & 62.23 & 10.41 &&& 38.21 & 64.28 & 25.14 &&& 33.33 & 64.32 & 14.70 \\
			& Experiment 3 & 25.74 & 25.03 & 49.45 &&& 9.02 & 13.54 & 50.54 &&& 25.37 & 25.63 & 49.45 &&& 13.26 & 17.56 & 48.51 \\
			& Experiment 4 & 25.33 & 62.54 & 26.92 &&& 9.02 & 63.11 & 35.71 &&& 25.08 & 64.34 & 26.92 &&& 13.26 & 64.31 & 18.57 \\
			\hline
			\multirow{4}{3cm}{Vanilla Multimodal AE}
			& Experiment 1 & 76.11 & 62.77 & 36.11 &&& 73.06 & 63.66 & 22.14 &&& 76.11 & 62.77 & 36.11 &&& 73.81 & 52.34 & 20.27 \\
			& Experiment 2 & 46.11 & 63.88 & 20.55 &&& 44.34 & 61.51 & 25.71 &&& 44.11 & 63.88 & 20.55 &&& 40.40 & 60.16 & 22.70 \\
			& Experiment 3 & 27.77 & 44.44 & 51.11 &&& 21.66 & 37.27 & 52.53 &&& 27.77 & 44.44 & 51.11 &&& 22.43 & 38.23 & 51.11 \\
			& Experiment 4 & 32.22 & 72.77 & 62.22 &&& 38.95 & 66.33 & 62.32 &&& 32.22 & 72.77 & 62.22 &&& 28.87 & 66.44 & 60.58 \\
			\hline
			\multirow{4}{3cm}{CorrNet}
			& Experiment 1 & 68.65 & 67.70 & 57.51 &&& 76.83 & 66.74 & 53.26 &&& 68.65 & 67.70 & 57.51 &&& 70.55 & 66.86 & 53.55 \\
			& Experiment 2 & 59.13 & 73.47 & 54.03 &&& 71.05 & 74.51 & 51.25 &&& 59.13 & 73.47 & 54.03 &&& 58.38 & 73.63 & 51.97 \\
			& Experiment 3 & 58.36 & 64.04 & 61.35 &&& 56.60 & 73.01 & 60.30 &&& 58.36 & 64.04 & 61.35 &&& 55.00 & 65.73 & 60.48 \\
			& Experiment 4 & 48.26 & 73.37 & 66.63 &&& 61.65 & 73.53 & 66.32 &&& 48.26 & 73.37 & 66.63 &&& 47.16 & 73.38 & 66.41 \\
			\hline
			\multirow{4}{3cm}{\textbf{Proposed}}
			& Experiment 1 & \textbf{96.99} & \textbf{91.53} & \textbf{81.24} &&& \textbf{98.12} & \textbf{94.26} & \textbf{85.72} &&& \textbf{96.99} & \textbf{91.53} & \textbf{81.24} &&& \textbf{97.21} & \textbf{90.71} & \textbf{79.77} \\
			& Experiment 2 & \textbf{95.97} & \textbf{93.44} & \textbf{88.73} &&& \textbf{97.31} & \textbf{95.62} & \textbf{89.78} &&& \textbf{95.97} & \textbf{93.44} & \textbf{88.73} &&& \textbf{96.18} & \textbf{92.52} & \textbf{88.71} \\
			& Experiment 3 & \textbf{96.56} & \textbf{93.23} & \textbf{89.27} &&& \textbf{97.75} & \textbf{95.52} & \textbf{90.26} &&& \textbf{96.56} & \textbf{93.23} & \textbf{89.27} &&& \textbf{96.77} & \textbf{92.36} & \textbf{89.28} \\
			& Experiment 4 & \textbf{96.27} & \textbf{93.23} & \textbf{88.98} &&& \textbf{97.56} & \textbf{95.52} & \textbf{89.97} &&& \textbf{96.27} & \textbf{93.23} & \textbf{88.98} &&& \textbf{96.48} & \textbf{92.36} & \textbf{88.95} \\
			
			\bottomrule
			\label{Table2}
		\end{tabular}%
	}
\end{table*}
\paragraph{}
The vanilla multi-modal Autoencoder objective function forces the network to maximize the predictability of both modalities from all three join-modal, single-modal $v_1$, and single-modal $v_2$ input passing modes. The objective function of the contrastive vanilla multi-modal Autoencoder without missing modality only forces the network to maximize the predictability of the acoustic and vibration modalities given both and also forces the joint-modal latent representation of the samples within the same class to be similar and the representation of the samples belonging to different classes to be dissimilar. CorrNet improves vanilla multi-modal Autoencoder objective function by maximizing the correlation of the single-modal representation of the input $v_1$ and the single-modal representation of the input $v_2$, $h(v_1)$ and $h(v_2)$. The performance metrics of the contrastive vanilla multi-modal Autoencoder without missing modality drop drastically while one of the input modalities is omitted. Although vanilla multi-modal Autoencoder with missing modality and CorrNet attempt to alleviate this issue and hamper the performance reduction caused by omitting one of the input modalities, these algorithms fail to keep the performance high enough on the missing modality. The most remarkable performance is achieved by our proposed methodology.
\paragraph{}
It is noteworthy that a dataset with very limited samples (about $700$) is used for training and evaluating our proposed methodology, and each train split in each fold of the $7$-fold stratified cross-validation procedure contains only about $600$ samples, which is extremely small. Moreover, the selected case study for evaluating our proposed methodology is a very complex engineered mechanism, and for the simulation of faulty scenarios, a very slight fault is applied (see section \ref{section.intro} for more detail). The performance on such a small dataset with a sophisticated fault nature promises that the methodology can perform very well on other types of equipment as well. Moreover, using a larger dataset with sufficient samples can improve the performance much more.
\paragraph{}
In Figures \ref{fig9.cms}-\ref{fig12.cms}, the confusion matrices of the test split of an arbitrary fold of the $7$-fold stratified cross-validation procedure are depicted. Based on these Figures, apparently, one of the input modalities can be omitted during test time without any performance reduction at all or with extremely slight sacrifice in performance.

\begin{figure}[p]
	\centering
	\subfigure[Joint-modal input, acoustic reconstruction]{\includegraphics[width=0.45\textwidth]{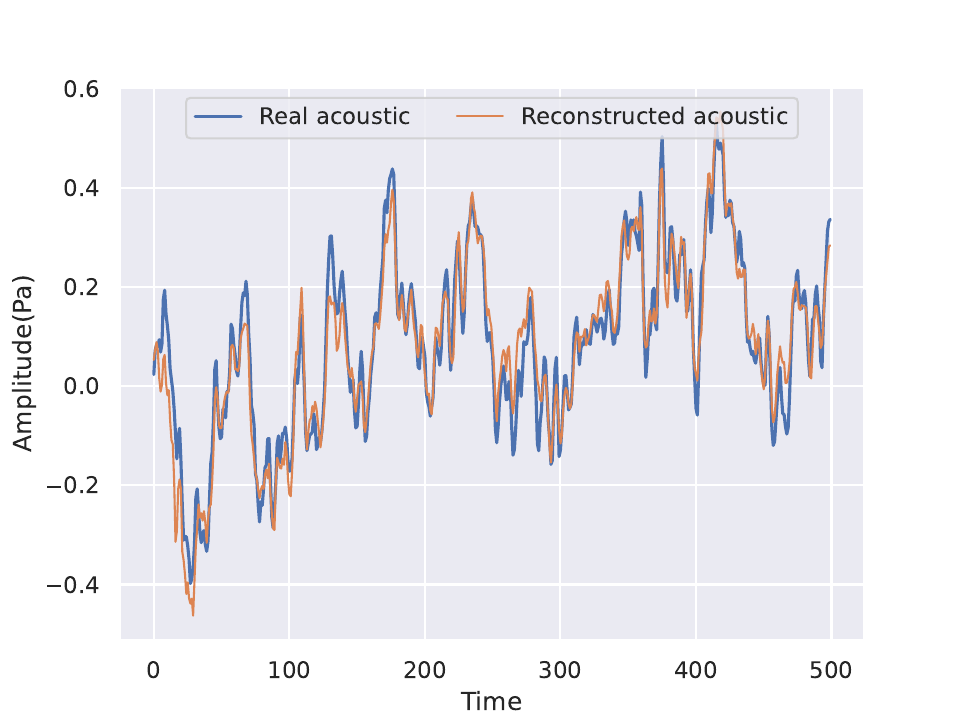}}
	\hfill
	\subfigure[Joint-modal input, vibration reconstruction]{\includegraphics[width=0.45\textwidth]{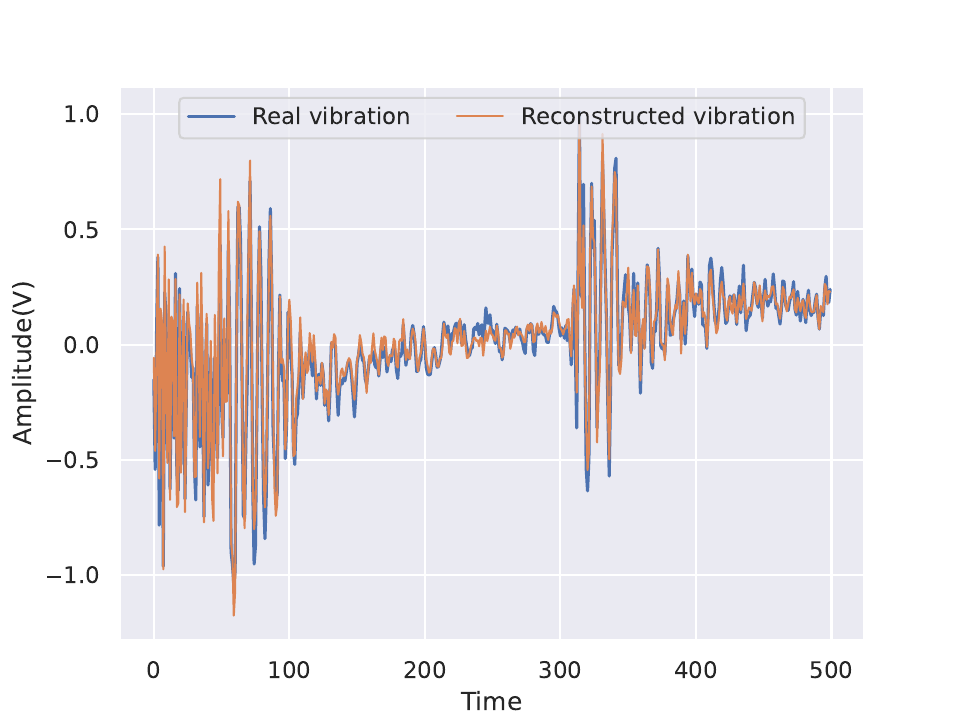}}
	\caption{Joint-modal input}\label{fig20.rec}
\end{figure}

\begin{figure}[p]
	\centering
	\subfigure[Acoustic modality reconstruction from acoustic]{\includegraphics[width=0.45\textwidth]{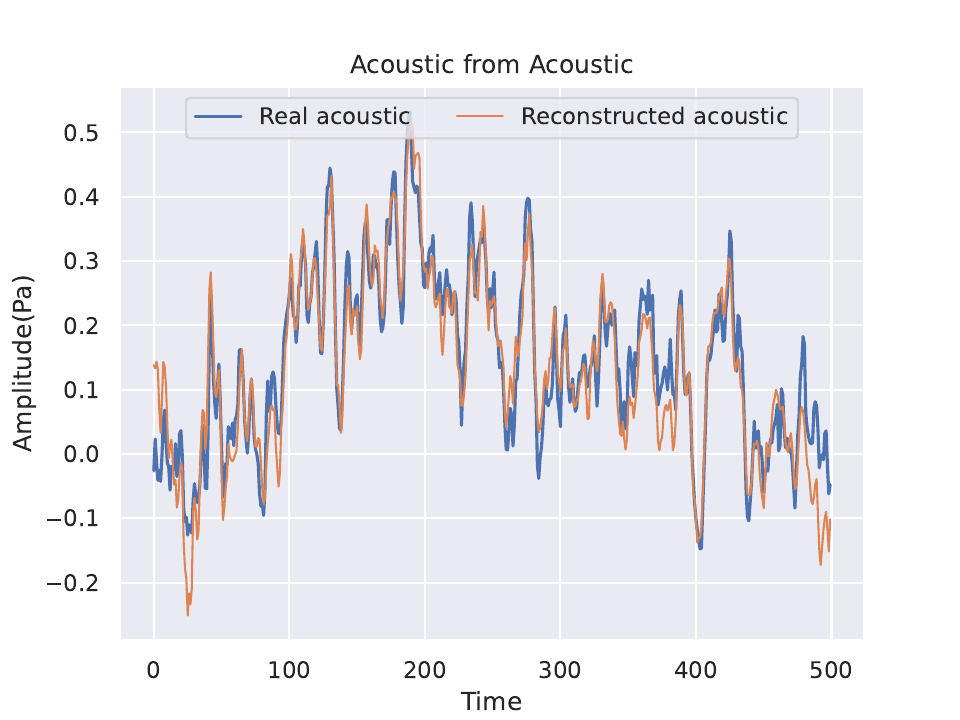}}
	\hfill
	\subfigure[Vibration modality reconstruction from acoustic]{\includegraphics[width=0.45\textwidth]{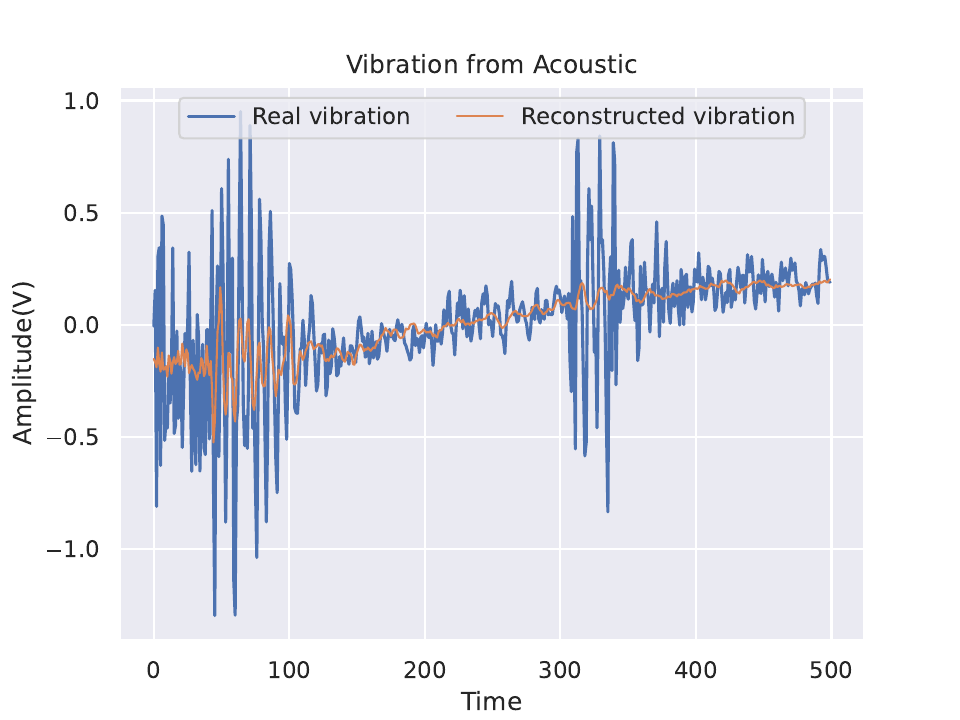}}
	\caption{Acoustic single-modal input}\label{fig21.rec}
\end{figure}

\begin{figure}[p]
	\centering
	\subfigure[Acoustic modality reconstruction from vibration]{\includegraphics[width=0.45\textwidth]{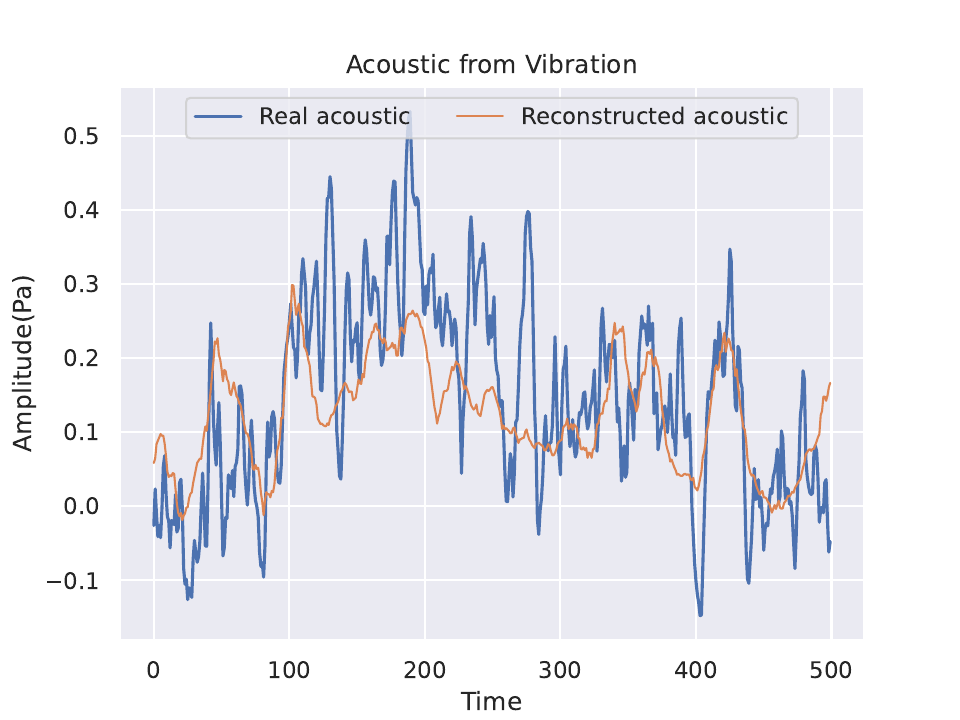}}
	\hfill
	\subfigure[Vibration modality reconstruction from vibration]{\includegraphics[width=0.45\textwidth]{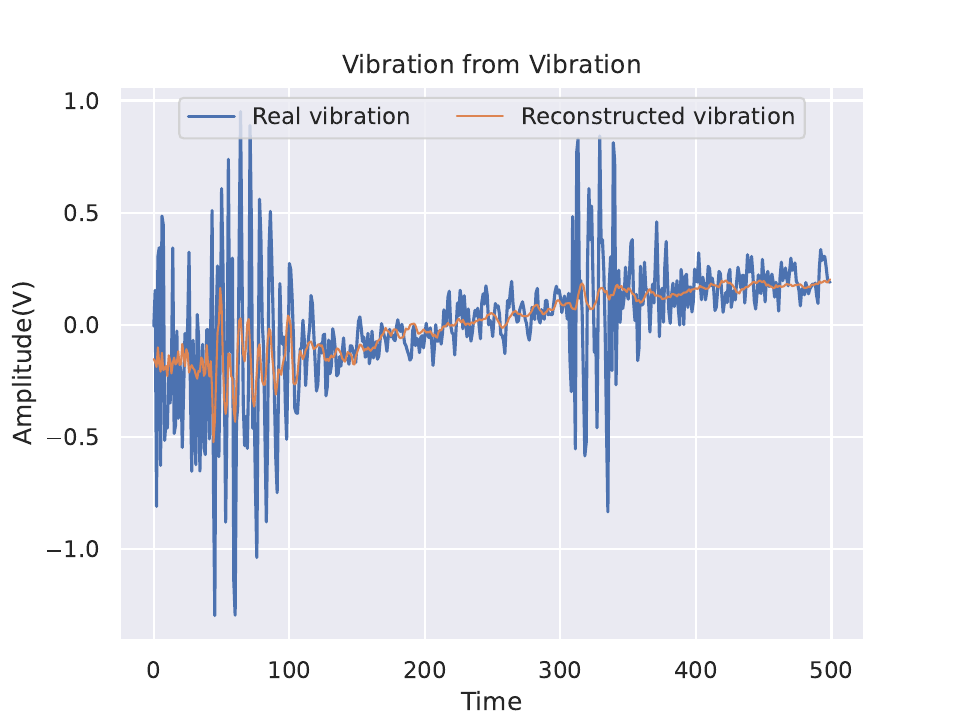}}
	\caption{Vibration single-modal input}\label{fig22.rec}
\end{figure}
\paragraph{}
For investigation of the reconstructed signals, a chunk of the reconstructed acoustic and vibration signals by the multi-modal Autoencoder in the three joint-modal, acoustic single-modal, and vibration single-modal input passing modes are depicted in Figures \ref{fig20.rec}-\ref{fig22.rec}. The reconstruction of the acoustic and vibration signals in joint-modal input passing mode in Figure \ref{fig20.rec} seems perfect by visual inspection. It can be seen in Figures \ref{fig21.rec} and \ref{fig22.rec} that in the missing modality case, the reconstruction of the present input modality is almost good, but cross-modality reconstruction (reconstruction of the missing modality) is not good and seems only contain the low frequency characteristics of the missing modality. This issue originates from the fact that we aim to reconstruct the missing modality only using the present modality features, which can only result in modality-invariant features reconstruction. 

\begin{figure}[!h]
	\centering
	\subfigure[Acoustic low frequency components]{\includegraphics[width=0.45\textwidth]{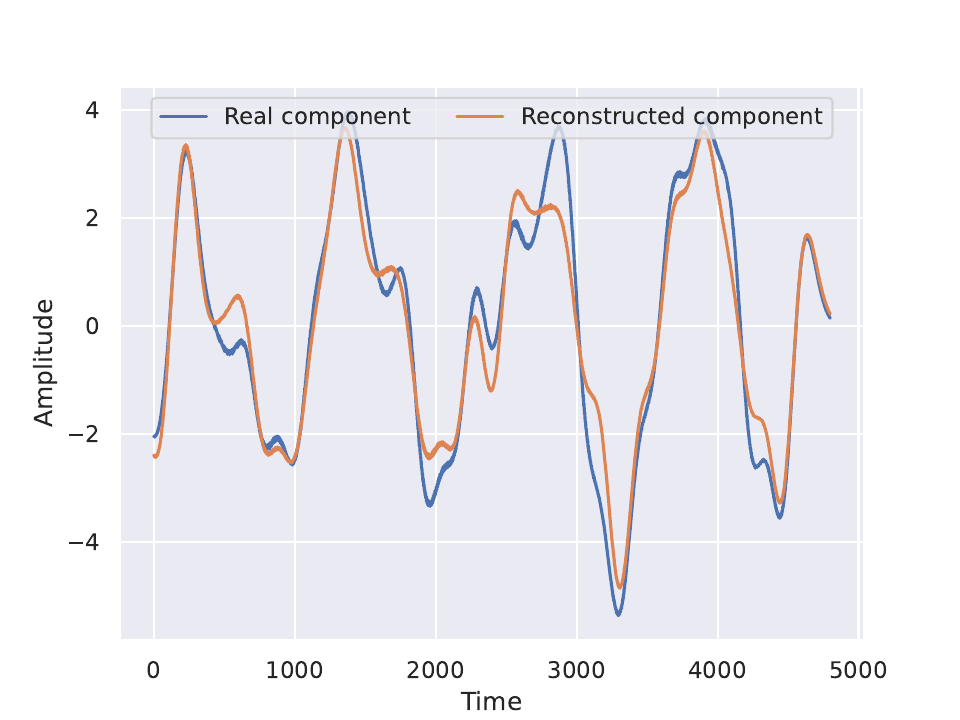}}
	\hfill
	\subfigure[Vibration low frequency components]{\includegraphics[width=0.45\textwidth]{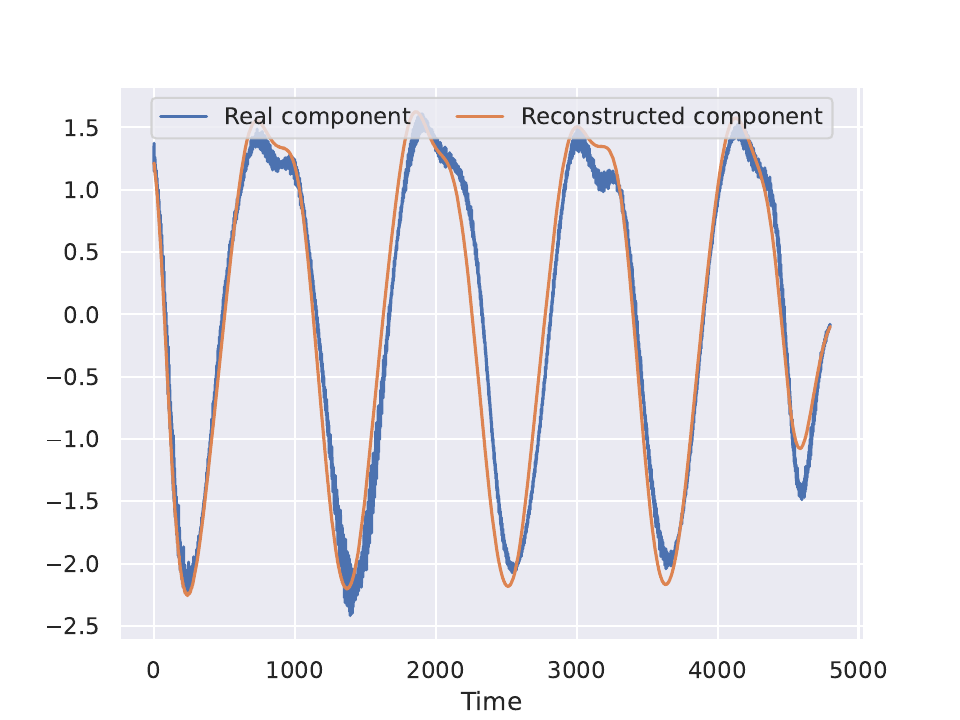}}
	\caption{Low frequency components, acoustic single-modal input}\label{fig21.wl}
\end{figure}

\begin{figure}[!h]
	\centering
	\subfigure[Acoustic low frequency components]{\includegraphics[width=0.45\textwidth]{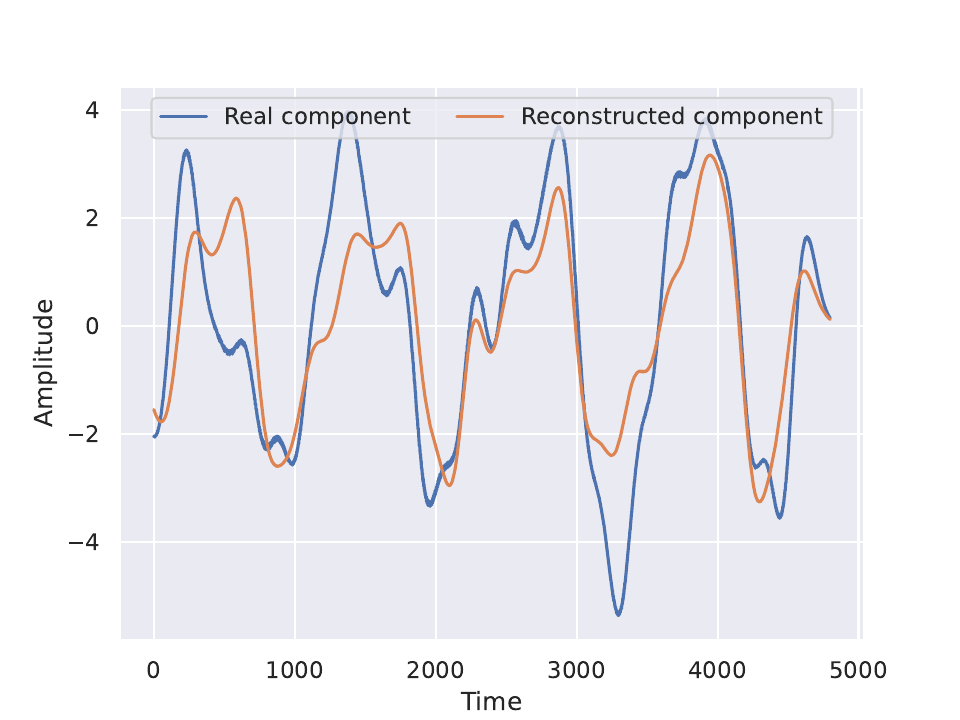}}
	\hfill
	\subfigure[Vibration low frequency components]{\includegraphics[width=0.45\textwidth]{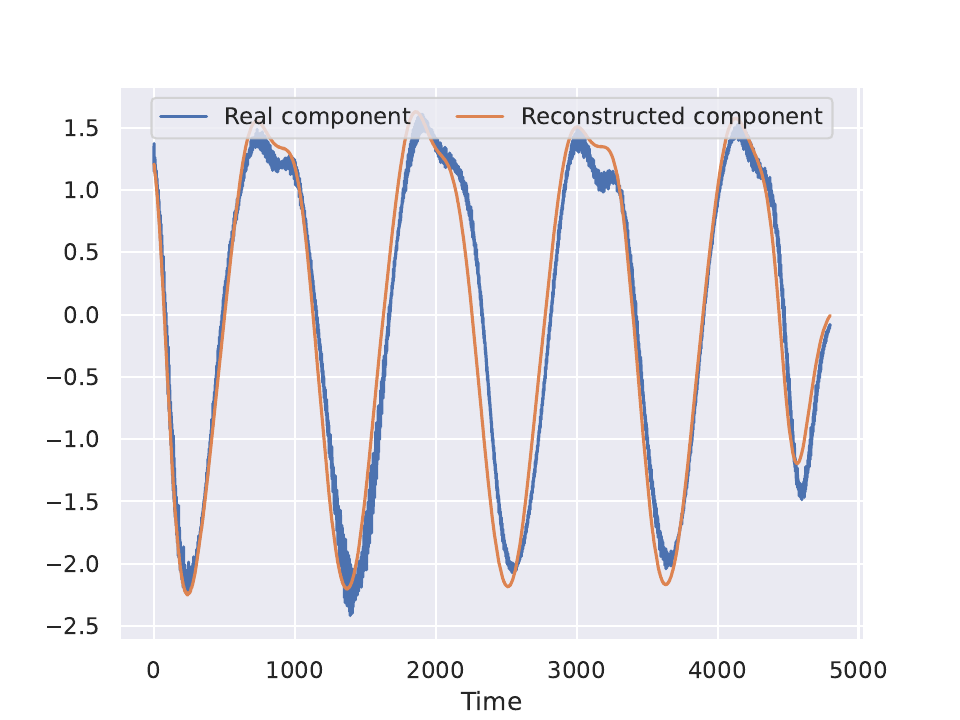}}
	\caption{Low frequency components, vibration single-modal input}\label{fig22.wl}
\end{figure}

Obviously, there are some modality-specific details that can not be perfectly reconstructed using only the common representation of the present modality, which leads to poor missing modality reconstruction.
\paragraph{}
The same issue exists in the previous works in \cite{22, 23, 24}. Moreover, it is clearly stated in \cite{23} that the perfect reconstruction of the missing modality is impossible when working with high-dimensional data. We utilized wavelet analysis \cite{57} to compare the real signals of each modality to the reconstructed ones. 
\paragraph{}
We used continuous Gaussian wavelet transform with $128$ decomposition levels and we figured out that the reconstructed signals contain the low frequency components of the real signals, and the error between the real and the reconstructed signals is due to their difference in high frequency components, which are originated from modality-specific details and variations. The low frequency components of the real and reconstructed acoustic and vibration signals for single-modal acoustic and vibration inputs are depicted in Figures \ref{fig21.wl} and \ref{fig22.wl} respectively.

\begin{figure}[h!]
	\centering
	\subfigure[Acoustic high frequency components]{\includegraphics[width=0.45\textwidth]{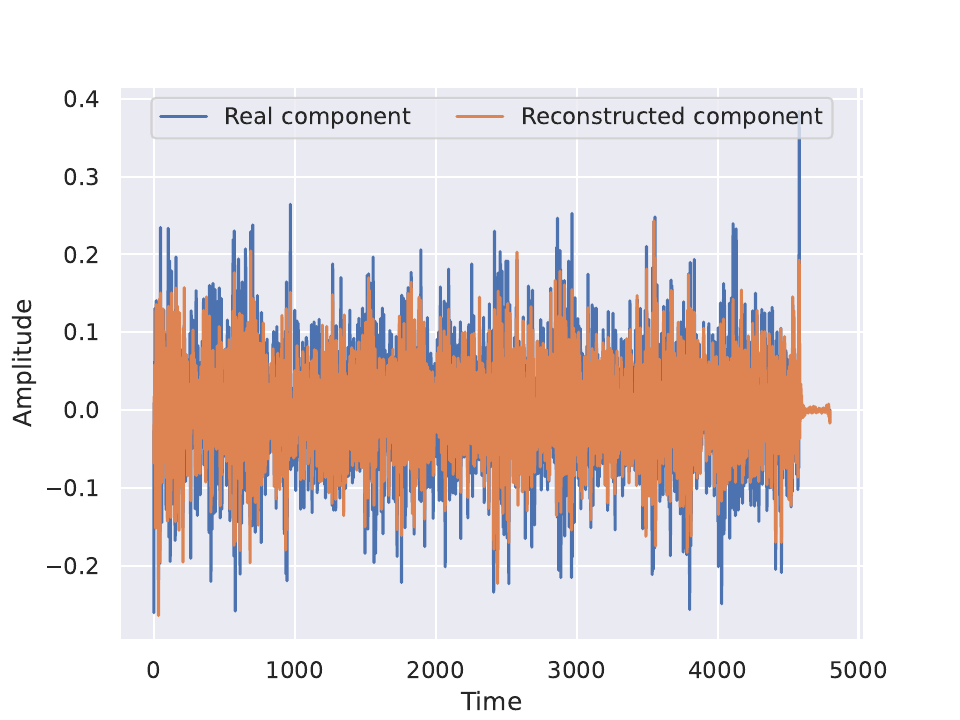}}
	\hfill
	\subfigure[Vibration high frequency components]{\includegraphics[width=0.45\textwidth]{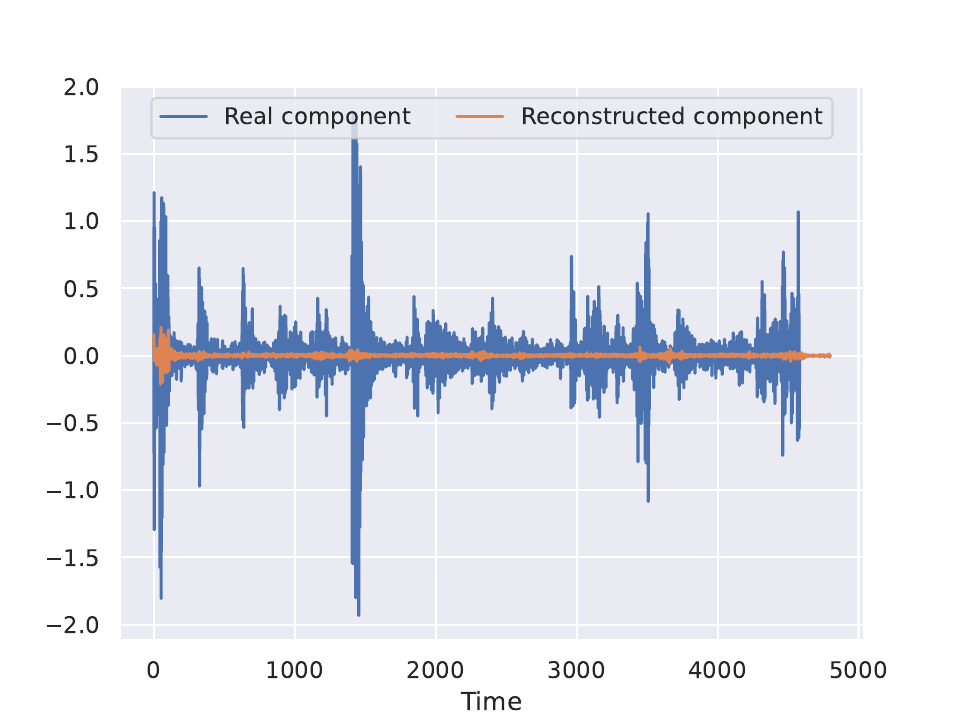}}
	\caption{High frequency components, acoustic single-modal input}\label{fig23.wl}
\end{figure}

\begin{figure}[h!]
	\centering
	\subfigure[Acoustic high frequency components]{\includegraphics[width=0.45\textwidth]{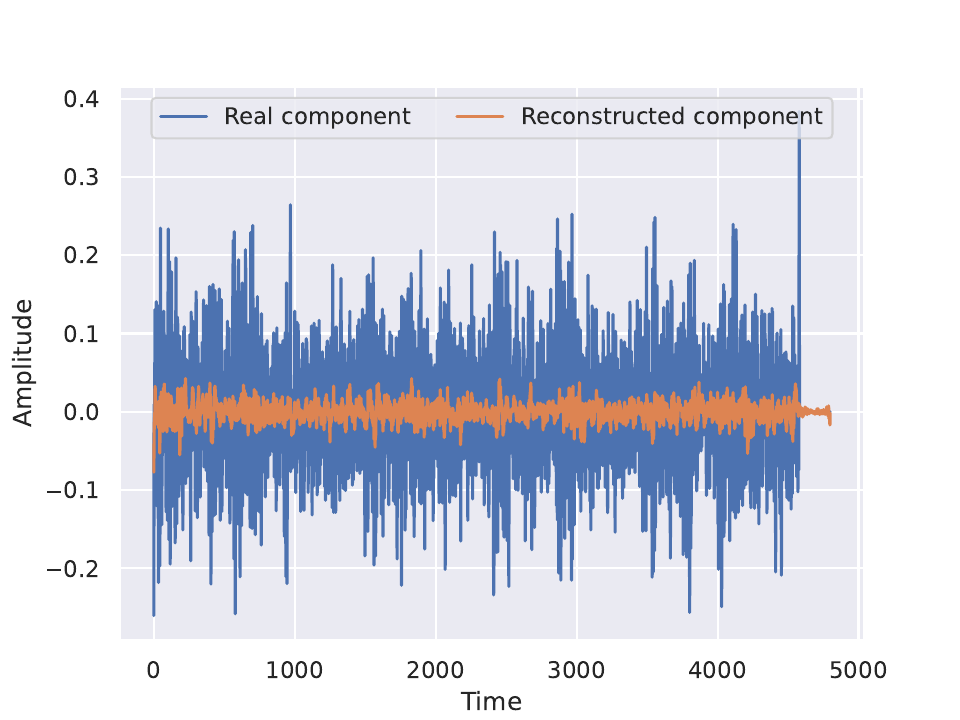}}
	\hfill
	\subfigure[Vibration high frequency components]{\includegraphics[width=0.45\textwidth]{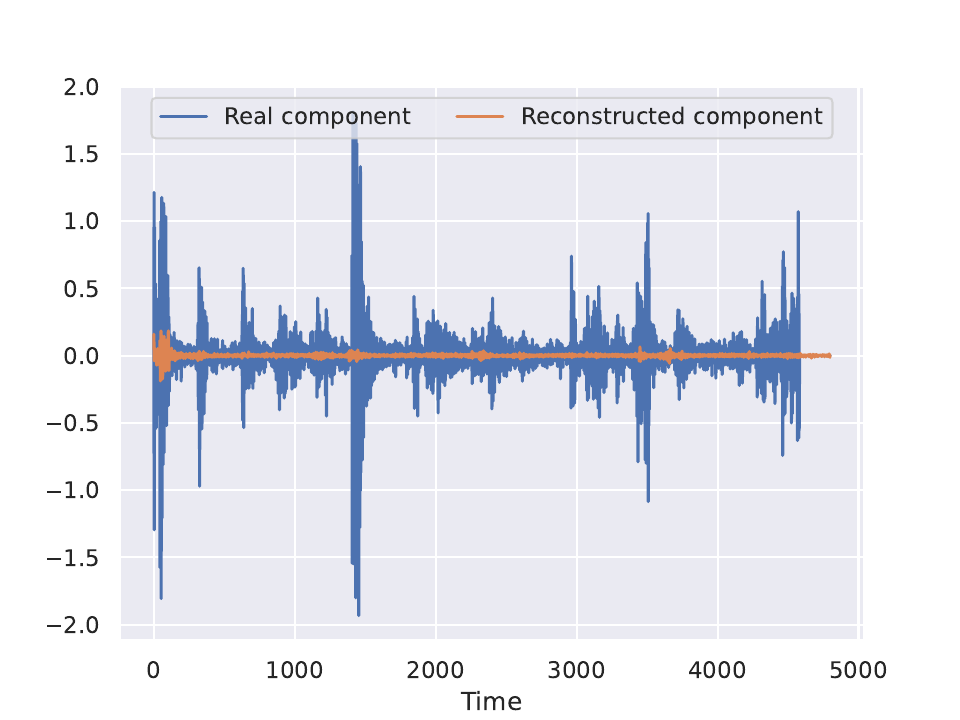}}
	\caption{High frequency components, vibration single-modal input}\label{fig24.wl}
\end{figure}

\paragraph{}
It is observed that the low frequency components of the real and the reconstructed signals are almost identical. The high frequency components of the real and the reconstructed acoustic and vibration signals in single-modal acoustic and vibration input passing modes are depicted in Figures \ref{fig23.wl} and \ref{fig24.wl} respectively. It is observed that the high frequency components of the reconstructed signals are not close to the high frequency components of real signals and that is the difference between real and reconstructed signals. Notably, the decoders role is to warrant that each of $h(z)$, $h(v_1)$, and $h(v_2)$ representations has a good essence or semantic of both of the input modalities, and since the downstream task of the proposed methodology is classification, the decoders are omitted during inference time and only the encoders and the fusion layer are kept. As a result, the imperfect reconstructions are negligible and have no adverse effect on the aim of current research, which is equipment health state classification.

\paragraph{}
In other words, the proposed methodology is performing data fusion and dimensionality reduction simultaneously through multi-modal representation learning with the aim of classification task, so perfect missing modality reconstruction is not required at all. Note that imperfect reconstruction and lack of high frequency contents in the reconstruction of missing modality does not mean that high frequency features are not extracted at all. On the contrary, very informative high and low frequency features are extracted by encoders since joint-modal reconstructions depicted in Figure \ref{fig20.rec} are almost perfect and include both high and low frequency components and classification performance on the latent representation depicted in Table \ref{Table1} is very good. Imperfect reconstruction of the missing modality implies that in missing modality case, the decoders are not able to combine the features extracted by the present view encoder to reconstruct all high frequency components (modality-specific detail) of the missing modality and are only capable of approximate reconstruction of it, which depicts that the correlations and connections between vibration and acoustic modalities are extracted in the latent space, and the common representation has an essence of both modalities.
\paragraph{}
Mathematical modeling of the reason behind the imperfect reconstruction of the missing modality and improving the quality of missing modality reconstruction are held for future works.
\section{Conclusion} \label{section.conclusion}
\paragraph{}	
In this research, we presented a contrastive multi-modal Autoencoder architecture based on Multi-View Learning paradigm with a novel training strategy that incorporates feature extraction, dimensionality reduction, and data fusion into a single process for multi-modal data aiming for intelligent condition monitoring and fault diagnosis. The proposed method not only achieves excellent performance in learning an enriched common representation from multi-modal sensory measurements and combines multi-modal data very well but also takes data fusion a further step wherein one of the input modalities can be accidentally or deliberately omitted during the inference time with a slight sacrifice in fault diagnosis performance or none at all. As a result, a multi-modal fault diagnosis system developed based on our proposed methodology achieves much higher robustness in case of sensor failure occurrence. Also, one can intentionally omit the more expensive sensor or the one that is harder to install in order to achieve a more cost-effective or an easy-to-use intelligent condition monitoring system for production line. Experimental results on a complex engineered case study with slight faults approve the capability of the proposed methodology in achieving the above-mentioned goals. Specifically, we demonstrated that our proposed method achieves higher performance than a wide variety of other methods, such as vanilla multi-modal Autoencoder and CorrNet, while one of the modalities is omitted during the test time. The main reason for the superiority of the proposed method over other methods is the existence of various contrastive losses, which force the network to extract and learn correlations and connections between different modalities, specifically vibration and acoustic modalities in current research and project different input modalities to a common space which includes discriminative features of these modalities. In this regard, t-sne visualization algorithm was used to demonstrate the latent representation of different samples, which shows different representations of samples belonging to the same class are gathered together and form dense cluster in the latent space far from representation clusters of other classes. Since multi-modal data and appropriate fusion strategy play an important role in advancing the performance level of health state classification, we believe our contribution is very beneficial since it can increase the performance of the fault diagnosis system without the additional cost of adding a real sensor during the inference time. Extending the methodology to cases with more than two modalities, mathematical modeling of the reason behind the imperfect reconstruction of the missing modality, and improving the missing modality reconstruction quality are held for future works.								 							
\end{document}